\documentclass[journal]{IEEEtran}
\usepackage{changes}

\usepackage{amsmath,epsfig}
\usepackage{pbox}
\usepackage{tikz}
\usetikzlibrary{arrows.meta,calc,decorations.markings,math,arrows.meta,positioning}
\usepackage{pgfgantt}
\usepackage{subfigure}
\usepackage{gensymb}
\usepackage{amsmath}
\usepackage{adjustbox} 
\usepackage{amsfonts}

\usepackage{hyperref}
\usepackage{multirow}
\usepackage{todonotes}

\usepackage{array}
\usepackage{graphicx}
\usepackage{multirow}

\usepackage{algorithm}
\usepackage{algpseudocode}

\DeclareMathOperator*{\argmax}{arg\,max}

\usepackage{soul}

\ifCLASSINFOpdf
\else
\fi
\usepackage{pbox}
\usepackage{tikz}
\usepackage{pgfgantt}
\usepackage{subfigure}
\usepackage{gensymb}
\usepackage{amsmath}

\usepackage{array}
\usepackage{graphicx}
\usepackage{multirow}
\usepackage{xcolor,colortbl}
\usepackage[normalem]{ulem}

\usetikzlibrary{matrix,shapes,arrows,positioning,chains}
\tikzstyle{process} = [rectangle, minimum width=3cm, minimum height=1cm, text centered, draw=black]
\tikzstyle{arrow} = [thick,->,>=stealth]
\tikzstyle{io} = [trapezium, trapezium left angle=70, trapezium right angle=110, text centered]

\makeatletter
\def\@copyrightspace{\relax}
\makeatother

\begin{document}
%
\title{Self-Supervised Multisensor Change Detection}
%
%
%

\author{
        Sudipan~Saha,~\IEEEmembership{Member,~IEEE,}
        Patrick~Ebel,
        and~Xiao~Xiang~Zhu,~\IEEEmembership{Fellow,~IEEE}
\thanks{Sudipan Saha is with Data Science in Earth Observation, Technical University of Munich, Taufkirchen/Ottobrunn, Germany. E-mail: sudipan.saha@tum.de  }
\thanks{Patrick Ebel is with Data Science in Earth Observation, Technical University of Munich, Taufkirchen/Ottobrunn, Germany. }
\thanks{Xiao Xiang Zhu is with Remote Sensing Technology Institute, German Aerospace Center (DLR), We{\ss}ling, Germany and also with Data Science in Earth Observation, Technical University of Munich, Taufkirchen/Ottobrunn, Germany. }
\thanks{Published in IEEE Transactions on Geoscience and Remote Sensing. Please visit the journal's page for the official version.}}


\maketitle

\begin{abstract}
Most change detection methods assume that pre-change and post-change images are acquired by the same sensor. However, in many real-life scenarios, e.g., natural disaster, it is more practical to use the latest available images 
before and after the occurrence of incidence, which may be acquired using different sensors. In particular, we are interested in the combination of the images acquired by optical and Synthetic 
Aperture Radar (SAR) sensors. SAR images appear vastly different from the optical images even when capturing the same scene.
Adding to this, change detection methods are often constrained to use only target image-pair, no labeled data, and no additional unlabeled data. Such constraints limit the scope of traditional
supervised machine learning and unsupervised generative approaches for multi-sensor change detection. Recent rapid development of self-supervised learning methods has shown that some of them can even work with
only few images. Motivated by this, in this work we propose a method for multi-sensor change detection using only the unlabeled target bi-temporal images that
are used for training a network in self-supervised fashion by using deep clustering and contrastive learning. The proposed method is evaluated on four multi-modal bi-temporal scenes showing change and the benefits of 
our self-supervised approach are demonstrated.

\end{abstract}


\fboxsep=0mm
\fboxrule=0.1pt

%
\IEEEpeerreviewmaketitle

\bstctlcite{IEEEexample:BSTcontrol}

\section{Introduction}
\label{secIntro}
Our Earth is rapidly changing, both due to natural and man-made causes. Satellite image based change detection (CD)  is generally used to monitor the temporal evolution of the
dynamic Earth \cite{saha2019unsupervised,appice2019empowering,chen2019change,rahman2018siamese,pomente2018sentinel,seydi2017new,puhm2020near}. CD ingests bi-temporal images as input and segregates all pixels as changed/unchanged. CD is a crucial step for several applications, including disaster
management, urban monitoring, forestry, glacier monitoring, and precision agriculture. Considering
the variation of applications, rarity of occurrences of some change-inducing incidents (e.g., natural disasters), and large geographic variation, it is imprudent to assume that
large-scale training datasets corresponding to all such tasks can be ever collected. Thus, there is a significant inclination in the CD literature towards  methods that can process
the target bi-temporal region-of-interest without using any training label or any additional pool of unlabeled image. Motivated by its excellent performance in computer vision, researchers have applied deep learning
to satellite image change detection \cite{ball2017comprehensive}. To exploit the potential of deep learning while not using any training label or additional unlabeled images, transfer learning based CD methods are popular that 
reuse a pre-trained network for bi-temporal feature extraction and comparison \cite{saha2019unsupervised}.
\par
A striking feature of satellite data is its variability, in terms of different sensors. Images captured using passive optical sensor are quite similar to the natural images studied in the computer vision. However, images
captured by the active sensors, e.g., Synthetic Aperture Radar (SAR) are remarkably different from the optical images \cite{hirschmugl2020use,zhou2020geo,ahmed2020sar}. While optical sensors use wavelengths near to visible light (approx. 1 micron), SAR uses a
wavelength of 1 cm to 1 m. Moreover, optical sensors rely upon the  natural illumination (e.g., sun) to create the brightness observed by the sensor, while the SAR sensors
carry their own illumination source, in the form of radio waves transmitted by an antenna. Moreover, satellite images
are captured with different number of spectral bands (one to few hundreds), different
spatial resolutions (few cm/pixel to Kms/pixel), different polarization.  While this vast variation provides an opportunity for detailed Earth observation, it is not trivial to use same set of methods
for images from different sensors. Due to this reason, most existing CD methods assume that the pre-change and post-change
images are acquired using the same sensor. The temporal frequency at which same sensor can image same place depends on the revisit period of the satellite on which
the sensor is mounted. However, better the spatial resolution, more close the satellite
is to the Earth, more time it takes to revisit same place. This is a hindrance in the use of same-sensor CD in time-bound applications, e.g., fast response
for disaster management and precision agriculture. Using different sensors may allow us to obtain temporal sequences with better temporal frequency without sacrificing spatial resolution. However, it 
is not trivial to process multi-sensor bi-temporal images as they are affected by the spectral characteristics of the sensors. Moreover, different sensors capture different type of information, making their comparison
often challenging \cite{schmitt2016data}. The difficulty of this problem is further accentuated by the fact that we are interested to detect change without using any labeled 
training data or any abundant pool of unlabeled data. 
\par
\begin{figure}[!t]
\centering
\subfigure[]{%
            \includegraphics[height=3.5 cm]{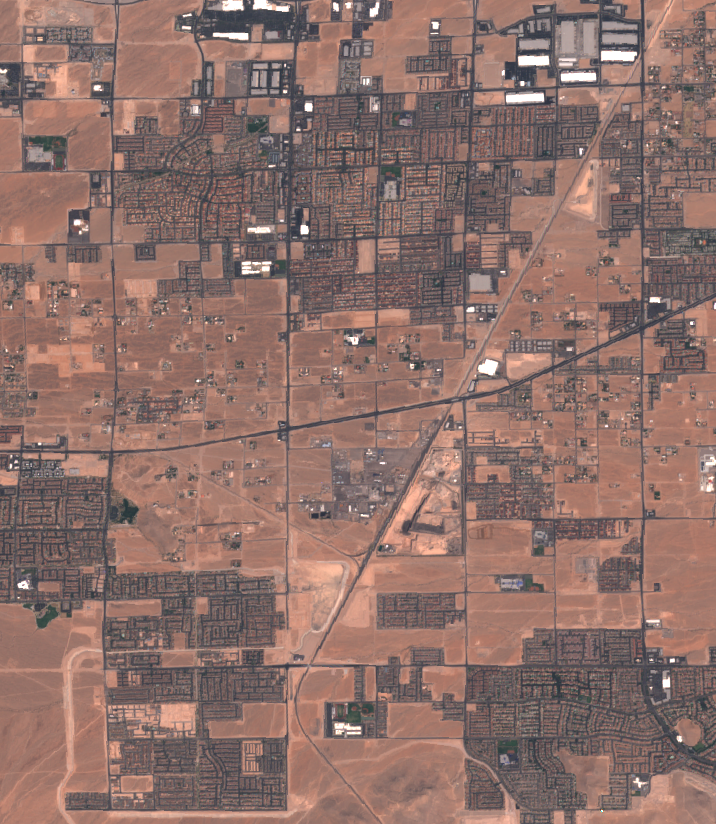}
            \label{inputImageWorldviewPre}
        }%
\subfigure[]{%
            \includegraphics[height=3.5 cm]{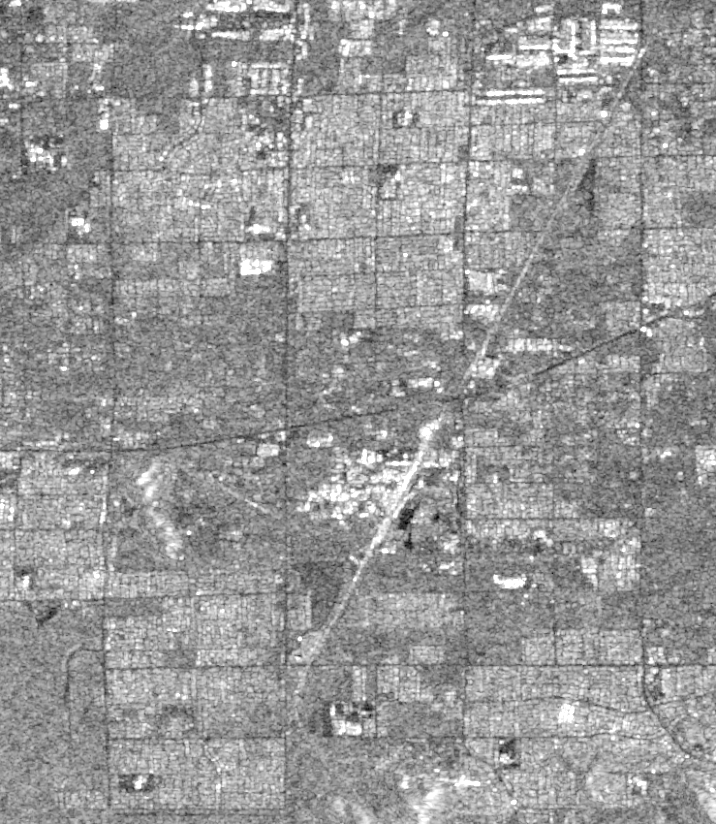}
            \label{inputImageWorldviewPost}
        }

\caption{Visual contrast for Las Vegas between (a) optical image (prechange) and (b) SAR image (postchange) . Optical and SAR images emphasize different properties of the target area, thus performing CD on them is challenging.}
\label{figureVisualContrastCityLasVegas}
\vspace{-0.4cm}
\end{figure}

The emergence of deep learning has seen
many such problems solved that were thought to be very challenging in the past \cite{zhu2017deep,camps2021deep}. Self-supervised learning has shown remarkable success recently, even when
only few images are available \cite{asano2019critical}.
Intrigued by this, in this paper we explore the challenging problem of change detection between optical and SAR images, the 
disparity between which is evident in Figure \ref{figureVisualContrastCityLasVegas}. We
exploit recent developments in the self-supervised
learning and deep clustering to propose a method for challenging SAR-optical CD where one of the bi-temporal images is acquired by an optical sensor, while the other is acquired by a SAR sensor.
\par
The proposed method requires only the bi-temporal target scene (where change is to be detected), no 
training label, and no additional unlabeled data. The target bi-temporal scene is typically large, few hundred pixels by few hundred pixels. Smaller bi-temporal patches (e.g., $64\times64$) are extracted from it
to train a two-branch network, similar to the Siamese network \cite{chen2020exploring}. Each branch of the network has a projection module and a predictor. Projection modules 
learn features unique to optical and SAR data without sharing weights, while predictors share the weight. The output of the predictors are used to estimate deep clustering loss for both images separately.
Moreover, considering prior probability of changed pixels are much less than the unchanged ones, a temporal consistency loss is proposed that ensures that pixels in the same location at two different time tend to 
get same label. To ensure that this does not lead the network to learn a trivial solution, a contrastive loss is used. By combination of these losses, the proposed method learns useful semantic feature from the multi-sensor (SAR-optical)
bi-temporal target scene and after training the network predictions can be compared for change detection.
\par
The contributions of this paper are as follows:
\begin{enumerate}
\item We propose a self-supervised learning method for change detection in a bi-temporal scene where one image is captured by the optical sensor and the other by the SAR sensor. The proposed method, only exploiting the available target 
unlabeled scene, effectively absorbs several concepts from the recent self-supervised learning literature, e.g., deep clustering, augmented view, Siamese network, and contrastive learning. By effectively
exploiting these concepts and modifying them appropriately for the
target multi-sensor bi-temporal data, proposed method is able to train a network that is further used for bi-temporal comparison and change detection.
\item We show the versatility of the self-supervised learning on spatio-temporal satellite data that is very different from typical computer vision images. Even though some form of aerial images (e.g., drone images) are often studied in the computer vision, 
we stress that our satellite data (both optical and SAR) are significantly different from the typical aerial images. 
\item We experimentally show the efficacy of the proposed method on four different bi-temporal multi-sensor scenes.
\end{enumerate}
\par
The rest of this paper is organized as follows. Related works are briefly discussed in Section \ref{sectionRelatedWork}. Section \ref{sectionProposedMethod} outlines the proposed method.
Datasets and experimental results are detailed in Section \ref{sectionExperimentalResult}. Finally, we conclude the paper in Section \ref{sectionConclusion}.

\section{Related Work}
\label{sectionRelatedWork}
In this Section we briefly discuss existing works on unsupervised CD (with focus on the multi-sensor CD) and self-supervised learning.

\subsection{Change Detection}
\label{subsectionCDRelatedWork}
Prior to the emergence of deep learning, most unsupervised change detection methods used the concept of pixelwise image differencing, i.e., change vector analysis (CVA) \cite{malila1980change}.
A number of superpixel and spatial neighborhood based variants of CVA have been proposed, e.g., Parcel Change Vector Analysis (PCVA) \cite{bovolo2009multilevel} and Robust Change Vector Analsis (RCVA) \cite{thonfeld2016robust}.
Most deep learning based unsupervised change detection methods use transfer learning. \cite{saha2019unsupervised} proposed deep change vector analysis (DCVA),
a CD framework that combines ideas from CVA with feature extraction based on pre-trained neural networks. In nutshell, a deep model that has been trained 
for some other task is reused to obtain pixelwise bi-temporal deep features from the target scene. Bi-temporal deep features
 are then compared to obtain deep change hypervectors for each pixel in the scene that
are analyzed based on magnitude ($\ell_2$ norm) to identify the changed pixels.
While \cite{saha2020building} shows that sensor-specific pre-trained network is more suitable for transfer learning,
\cite{pomente2018sentinel} advocates models trained on ImageNet \cite{deng2009imagenet} for transfer learning in CD. 
There is another class of unsupervised CD methods that pre-classifies some pixels with high confidence as changed/unchanged using 
some traditional approach and further uses those confident samples for training a CD model \cite{keshk2019change}.
\par
It is not trivial to process multi-sensor bi-temporal images as they are affected by differences in the spatial resolution and differences in
the spectral characteristics of the sensors. Due to this, there are very few works that can work in the setting where
pre-change and post-change images have different spatial resolution \cite{saha2019unsupervisedMultisensor, zhang2016change} or bands with
different spectral characteristics \cite{volpi2015spectral}. Moreover, those works deal with only minor variation in spatial or
spectral characteristics. \cite{saha2019unsupervisedMultisensor} proposed a cycle-consistent generative adversarial network based
method to learn transcoding between multi-sensor multi-temporal domain. However, their work assumes that 
a large (unlabeled) area corresponding to both sensors are available as training data. \cite{liu2016deep} used a symmetric convolutional
coupling network (SCCN) and \cite{zhan2018log} used denoising autoencoder (DAE)  for CD in multisensor  images. Though those works considered
optical-SAR images, they applied their methods on scenes with limited spatial complexity.
While our work is strongly motivated from the existing works on multi-sensor CD \cite{saha2019unsupervisedMultisensor, zhang2016change}, it takes them
a step further by considering the challenging scenario of optical-SAR CD in complex urban scenes and furthermore by integrating recent developments in self-supervised learning.

\subsection{Self-supervised learning}
\label{subsectionSelfSupLearningRelatedWork}
Considering the difficulty of collecting labeled data and abundance of unlabeled data, machine learning researchers have focused on developing unsupervised and self-supervised deep learning methods in the recent past.
Gidaris \textit{et. al.} \cite{gidaris2018unsupervised} used image rotation as a pre-text task to learn unsupervised semantic feature. Several other pre-text tasks have been
explored in the literature, e.g., relative patch prediction \cite{doersch2015unsupervised} and image inpainting \cite{pathak2016context}. 
Deep clustering, i.e., joint learning of the parameters of deep network and the cluster assignment of the resulting features,
has also been shown to be effective for unsupervised representation learning \cite{caron2018deep}. Remarkably, \cite{asano2019critical} have shown
that above-mentioned unsupervised methods learn useful semantic features even with a single-image
input. Contrastive methods function by bringing the representation of different views of the same image (‘positive pairs’) closer while spreading
representations of different images (‘negative pairs’) apart \cite{chen2020simple, chen2020improved, tian2020makes}. Boostrap your own latent \cite{grill2020bootstrap}
and its variant SiamSiam \cite{chen2020exploring} eliminate the requirement of negative pair by using multiple views of the same image. 
In more details, SiamSiam \cite{chen2020exploring} ingests as input two randomly augmented views of an image and processes it through a Siamese architecture. Each Siamese branch consists of
an encoder and a prediction head. The encoders share weight between two views. 
\par
\textit{Proposed method} is strongly inspired from the above self-supervised methods. Like deep clustering \cite{caron2018deep}, the proposed method 
uses the concept of simultaneous representation learning and cluster/label
assignment. The bi-temporal images can be considered to be views of same scene, like SiamSiam \cite{chen2020exploring}. Like the contrastive methods, the
proposed method uses the idea of bringing closer the representation 
of positive pairs and spreading apart the negative pairs.
Like \cite{asano2019critical}, the proposed method works on single scene (a pair of images capturing same location at two different times).
\par
\textit{Multi-temporal satellite image processing} researchers have also proposed self-supervised representation learning methods, e.g., deep clustering for multi-temporal segmentation \cite{saha2020unsupervised} and 
learning by rearranging randomly shuffled time-series images \cite{saha2020change}. Proposed method is related to them, using the concept of deep clustering as in \cite{saha2020unsupervised}.

\tikzset{
    between/.style args={#1 and #2}{
         at = ($(#1)!0.5!(#2)$)
    }
}
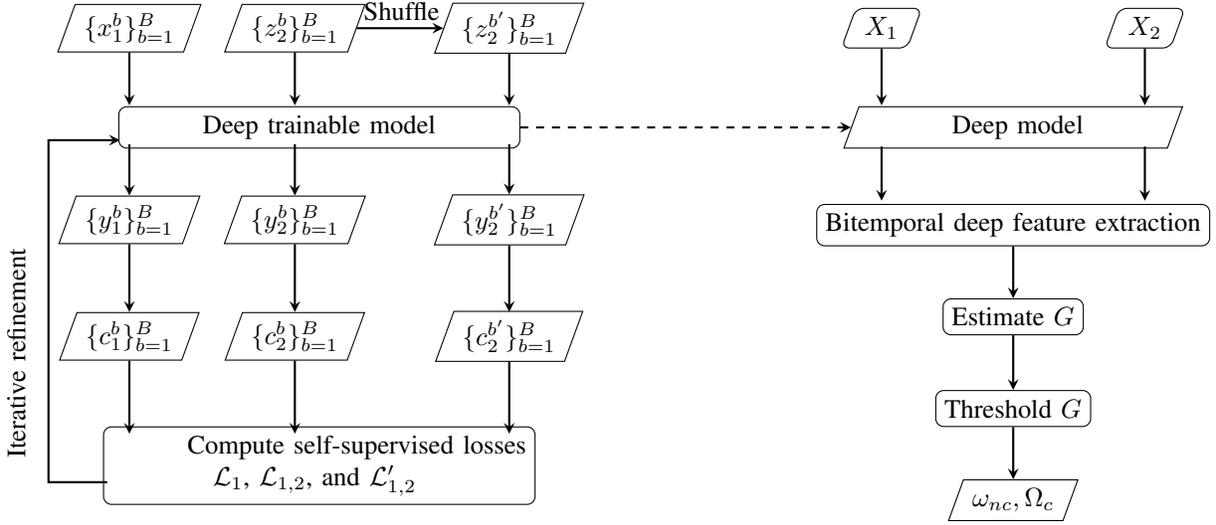
\begin{figure*}
 \centering
 \begin{tikzpicture}
 \node (1a) [draw, io, align=center] {$\{x^b_{1}\}_{b=1}^B$};
\node (1b) [draw, io, align=center, right of=1a,xshift=1.2cm] {$\{z^b_{2}\}_{b=1}^B$};
\node (1c) [draw, io,right of=1b,xshift=1.85cm] {$\{z^{b'}_{2}\}_{b=1}^B$};
\node (ancillaryText1) [right of=1b,xshift=0.42cm,yshift=0.22cm] {Shuffle};

\node (below1adot) [below of=1a,  yshift=-0.15cm] {};
\node (below1bdot) [below of=1b,  yshift=-0.15cm] {};
\node (below1cdot) [below of=1c,  yshift=-0.15cm] {};

\node (2) [rounded corners=3pt,draw, align=center, between=1a and 1c,  yshift=-1.32cm]{\hspace{1cm}Deep trainable model\hspace{1cm}};
\node (below2adot) [below of=1a,  yshift=-0.42cm] {};
\node (below2bdot) [below of=1b,  yshift=-0.42cm] {};
\node (below2cdot) [below of=1c,  yshift=-0.42cm] {};

 \node (3a) [io,draw, below of=1a,  yshift=-1.55cm] {$\{y^b_{1}\}_{b=1}^B$}; 
\node (3b) [io,draw, below of=1b,  yshift=-1.55cm] {$\{y^b_{2}\}_{b=1}^B$};
\node (3c) [io,draw, below of=1c,  yshift=-1.55cm] {$\{y^{b'}_{2}\}_{b=1}^B$};
\node (below3adot) [below of=3a,  yshift=-0.35cm] {};
\node (below3bdot) [below of=3b,  yshift=-0.35cm] {};
\node (below3cdot) [below of=3c,  yshift=-0.35cm] {};

 \node (4a) [io,draw, below of=3a,  yshift=-0.55cm] {$\{c^b_{1}\}_{b=1}^B$}; 
\node (4b) [io,draw, below of=3b,  yshift=-0.55cm] {$\{c^b_{2}\}_{b=1}^B$};
\node (4c) [io,draw, below of=3c,  yshift=-0.55cm] {$\{c^{b'}_{2}\}_{b=1}^B$};
\node (below4adot) [below of=4a,  yshift=-0.42cm] {};
\node (below4bdot) [below of=4b,  yshift=-0.42cm] {};
\node (below4cdot) [below of=4c,  yshift=-0.42cm] {};

\node (5) [rounded corners=3pt,draw, align=center, between=4a and 4c,  yshift=-1.72cm]{\hspace{1cm}Compute self-supervised losses \\ $\mathcal{L}_1$, $\mathcal{L}_{1,2}$, and $\mathcal{L}'_{1,2}$ \hspace{1cm}};
\node  [shape=coordinate] (leftOf5Point1) [left of=5,  yshift=-0.22cm, xshift=-1.82cm] {};
\node  [shape=coordinate] (leftOf5Point2) [left of=5,  yshift=-0.22cm, xshift=-2.6cm] {};
\node  [shape=coordinate] (leftOf2Point1) [left of=2,  yshift=-0.17cm, xshift=-2.6cm] {};
\node  [shape=coordinate] (leftOf2Point2) [left of=2,  yshift=-0.17cm, xshift=-1.65cm] {};
\node (ancillaryText2) [left of=4a, xshift=-0.50cm,yshift=-0.22cm,rotate=90] {Iterative refinement};

\draw [arrow] (1a) -- (below1adot);
\draw [arrow] (1b) -- (below1bdot);
\draw [arrow] (1b) -- (1c);
\draw [arrow] (1c) -- (below1cdot);

\draw [arrow] (below2adot) -- (3a);
\draw [arrow] (below2bdot) -- (3b);
\draw [arrow] (below2cdot) -- (3c);

\draw [arrow] (3a) -- (below3adot);
\draw [arrow] (3b) -- (below3bdot);
\draw [arrow] (3c) -- (below3cdot);

\draw [arrow] (4a) -- (below4adot);
\draw [arrow] (4b) -- (below4bdot);
\draw [arrow] (4c) -- (below4cdot);

\draw [arrow] (leftOf5Point1)--(leftOf5Point2)--(leftOf2Point1)--(leftOf2Point2);

 \node (1aTest) [rounded corners=3pt,draw, io, right of=1b,xshift=6.8cm] {$X_{1}$};
\node (1bTest) [rounded corners=3pt,draw, io, align=center, right of=1aTest,xshift=2.5cm] {$X_{2}$};
\node (below1adotTest) [below of=1aTest,  yshift=-0.15cm] {};
\node (below1bdotTest) [below of=1bTest,  yshift=-0.15cm] {};
\node (2Test) [io,draw, align=center, between=1aTest and 1bTest,  yshift=-1.32cm]{\hspace{1cm} Deep model\hspace{1cm}};
\node (below2adotTest) [below of=1aTest,  yshift=-0.45cm] {};
\node (below2bdotTest) [below of=1bTest,  yshift=-0.45cm] {};
\node (above3adotTest) [below of=1aTest,  yshift=-1.42cm] {};
\node (above3bdotTest) [below of=1bTest,  yshift=-1.42cm] {};

\node (4Test) [rounded corners=3pt,draw, align=center,  below of=2Test,  yshift=-0.30cm] {Bitemporal deep feature extraction};

\node (6Test) [rounded corners=3pt,draw, align=center, below of=4Test, yshift=-0.22cm] {Estimate $G$};
\node (7Test) [rounded corners=3pt,draw, align=center, below of=6Test, yshift=-0.22cm] {Threshold $G$};
\node (8Test) [io,draw, align=center, below of=7Test, yshift=-0.22cm] {$\omega_{nc}, \Omega_c$};

\draw [arrow] (1aTest) -- (below1adotTest);
\draw [arrow] (1bTest) -- (below1bdotTest);
\draw [arrow] (4Test) -- (6Test);
\draw [arrow] (6Test) -- (7Test);
\draw [arrow] (7Test) -- (8Test);

\draw [arrow] (below2adotTest) -- (above3adotTest);
\draw [arrow] (below2bdotTest) -- (above3bdotTest);

\draw [dashed,arrow] (2) -- (2Test);

\end{tikzpicture}
 \caption{Proposed unsupervised multi-sensor (optical-SAR) CD framework. The left hand side denotes the self-supervised training process while the right hand side shows the CD process using already trained model.}
 \label{unsupervisedMultisensorCDBlockDiagram}
\end{figure*}

\tikzset{
    between/.style args={#1 and #2}{
         at = ($(#1)!0.5!(#2)$)
    }
}
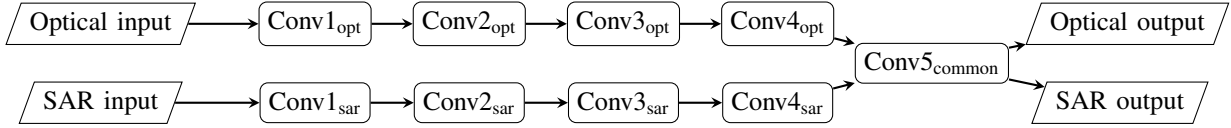
\begin{figure*}
 \centering
 \begin{tikzpicture}
 \node (1a) [draw, io, align=center] {Optical input};
 \node (1b) [draw,rounded corners=3pt,right of=1a,xshift=1.85cm] {Conv1\textsubscript{opt}};
 \node (1c) [draw, rounded corners=3pt,right of=1b,xshift=1.05cm] {Conv2\textsubscript{opt}};
  \node (1d) [draw, rounded corners=3pt,right of=1c,xshift=1.05cm] {Conv3\textsubscript{opt}};
   \node (1e) [draw, rounded corners=3pt,right of=1d,xshift=1.05cm] {Conv4\textsubscript{opt}};

\node (2a) [draw, io, align=center, below of=1a,  yshift=-0.05cm] {SAR input};
 \node (2b) [draw, rounded corners=3pt,right of=2a,xshift=1.85cm] {Conv1\textsubscript{sar}};
 \node (2c) [draw, rounded corners=3pt,right of=2b,xshift=1.05cm] {Conv2\textsubscript{sar}};
  \node (2d) [draw, rounded corners=3pt,right of=2c,xshift=1.05cm] {Conv3\textsubscript{sar}};
   \node (2e) [draw, rounded corners=3pt,right of=2d,xshift=1.05cm] {Conv4\textsubscript{sar}};

\node(3) [draw, rounded corners=3pt,between=1e and 2e,xshift=2.05cm] {Conv5\textsubscript{common}};

\node (4a) [draw, io,right of=1e,xshift=3.65cm] {Optical output};
\node (4b) [draw, io,right of=2e,xshift=3.55cm] {SAR output};

\draw [arrow] (1a) -- (1b);
\draw [arrow] (1b) -- (1c);
\draw [arrow] (1c) -- (1d);
\draw [arrow] (1d) -- (1e);

\draw [arrow] (2a) -- (2b);
\draw [arrow] (2b) -- (2c);
\draw [arrow] (2c) -- (2d);
\draw [arrow] (2d) -- (2e);

\draw [arrow] (1e) -- (3);
\draw [arrow] (2e) -- (3);

\draw [arrow] (3) -- (4a);
\draw [arrow] (3) -- (4b);

\end{tikzpicture}
 \caption{The network simplified architecture with $L_1=4$ and $L_2=1$. Optical and SAR inputs are processed separately and subsequently fed to a common prediction layer.}
 \label{figureNetworkArchitecture}
\end{figure*}

\section{Proposed method}
\label{sectionProposedMethod}
Let $X_{1}, Z_{2}$ be two images of size $R\times C$ taken over the same geographical region at time $t_1$ and $t_2$, respectively.
Without loss of generality we assume that the pre-change image $X_{1}$ is acquired by optical sensor (RGB) and the post-change image  $Z_{2}$ is
acquired by the SAR sensor.  Since SAR image is  grayscale, same channel is replicated thrice to make it 3-channel like the optical input.
We aim to detect changes from the images $X_{1}, Z_{2}$ in an unsupervised manner, i.e., without using any training labels
and any additional unlabeled data pool. Our goal is to divide the set of all pixels $\Omega$ into
two subsets $\Omega_{c}$ and $\omega_{nc}$ corresponding to changed and unchanged pixels, respectively. Like most existing
unsupervised CD methods \cite{saha2019unsupervised}, we assume that prior probability of occurrence of change is less compared to no change \cite{bovolo2015time}. 
\par
We can extract a set of bi-temporal patches of size $R'\times C'$ ($R'<R$ and $C'<C$) from the images $X_{1}, Z_{2}$.  
In practice, one training iteration involves only a batch of $\mathcal{B}$ patches from $X_{1}$, denoted as   $\mathcal{X}=\{x^1_{1},...,x^\mathcal{B}_{1}\}$ and corresponding patches from 
$Z_{2}$, denoted as  $\mathcal{Z}=\{ z^1_{2},...,z^\mathcal{B}_{2}\}$.  $x^b_{1}$ and $z^b_{2}$ are processed separately with deep clustering loss, as detailed in
Section \ref{sectionDeepClustering}. Furthermore, considering that  
$x^b_{1}$ and $z^b_{2}$ represent same location at two different times and prior probability of change is less, a temporal consistency loss (see Section \ref{spatioTemporalHomogeneity}) is
formulated using each such pair.
Furthermore, $\mathcal{Z}$ is shuffled to form negative samples $\mathcal{Z'}$ and a  contrastive loss is used between pairs from $\mathcal{X}$ and $\mathcal{Z'}$, as outlined in Section \ref{spatioConstrastiveLoss}.
Proposed method is outlined in Figure \ref{unsupervisedMultisensorCDBlockDiagram}.

\subsection{Bi-temporal patches are multiple views of the same location}
\label{subsectionMultisensorMultipleView}
We recall from Section \ref{subsectionSelfSupLearningRelatedWork} that many self-supervised learning approaches build upon the concept of bringing closer the representation of the
multiple views of the same image. Different views of the same image are generally obtained by different augmentation techniques, e.g., random crops.
We argue that multi-sensor bi-temporal patches $x^b_{1}$ and $z^b_{2}$ can be similarly thought to be multiple views of the same location. They represent  augmentation of the same place, where
the augmentation transformation is naturally caused by multisensor differences and other factors including weather condition. Considering that the prior probability of change is less \cite{bovolo2015time},  most of the times 
such pair of patches $x^b_{1}$ and $z^b_{2}$ represent the same information, but from the eyes of two different viewers (sensors).

\subsection{Siamese representation}
\label{subsectionSiameseRepresentation}
Since bi-temporal patches can be seen as multiple views of the same location, we argue that semantic information can be captured from them by using a Siamese-like architecture.
Similar to  \cite{chen2020exploring}, both branches of the two-branch network have projection modules $f_{opt}$ and $f_{sar}$ for the optical and SAR branch, respectively. Additionally, both branches have
prediction modules $h_{opt}$ and $h_{sar}$ for the optical and SAR branch, respectively.
However, unlike \cite{chen2020exploring}, 
the projection modules $f_{opt}$ and $f_{sar}$  do not share weight. This is because SAR and optical images are significantly different processed by two different projection
modules using different sets of weights. However, the prediction modules $h_{opt}$ and $h_{sar}$ share weights and henceforth simply denoted as $h$.
\par
The projection and the prediction networks consists of $L_1$ and $L_2$ (generally $L_2 = 1$) convolutional layers, respectively, where $L = L_1+L_2$.  The two projections
compute a projected representation from the optical and SAR images
and project them to a common domain. In the ideal scenario,
where the projectors have perfectly learned to project optical and SAR images into a common domain and the bi-temporal images do not show any change, the output generated for an input pair
is expected to be identical. However, practically even in absence of any change, there are differences caused by multi-sensor acquisition and other factors that are not trivial for projection modules to mitigate.
\par
All but last convolution layers are followed by ReLU activation function. They are further followed by batch normalization layer. We do not use any pooling layer, hence the
size of the input is preserved in the output. While filters of spatial size $3\times 3$ are used for all convolution layers for projection, the prediction module uses $1\times 1$ filter.
The kernel number of the final layer is $K$ and can be thought of as $K$ different clusters/classes. Each pixel can be assigned to one of these $K$ clusters (detailed in Section \ref{sectionDeepClustering}).
The network architecture is shown in Figure \ref{figureNetworkArchitecture}.

\subsection{Deep clustering}
\label{sectionDeepClustering}
Deep clustering process involves the joint learning of the parameters of deep network and the cluster assignment of the resulting features \cite{caron2018deep}.
Deep clustering helps the network to learn discriminative features that can identify different classes/clusters in the images. Considering processing of the two images as independent process, deep clustering can be performed for each of them.
The output obtained by the network for a paired input patches $x^b_{1}$ and $z^b_{2}$ is:
\begin{align}
y^b_1 = h(f_{opt}(x^b_{1}))\\
y^b_2 = h(f_{sar}(z^b_{2}))
\end{align}
$y^b_1$ has same spatial dimension $R'\times C'$ as $x^b_{1}$ and has kernel number (or, feature dimension) $K$. The deep clustering process is performed over the
pixels, i.e., each pixel is assigned to a cluster.
Without loss of generality, we henceforth explain the deep clustering process in reference to a generic pixel $y^b_{1,n}$ from $y^b_1$.
The dimension of $y^b_{1,n}$ is $K$ that can be converted to 1-dimensional label $c^b_{1,n}$ by argmax classification. 
This is achieved by selecting the kernel/feature in $y^b_{1,n}$ that has maximum value. If
 the $k$-th feature of $y^b_{1,n}$ is represented by $y^b_{1,n}(k)$, then label $c^b_{1,n}$ is obtained as following:
\begin{equation}
c^b_{1,n}=\argmax_{k\in K} y^b_{1,n}(k)
\end{equation}
The rationale behind finding the highest activation of an input pixel is that the pixels 
that obtain the highest activation in the same feature are likely to have similar semantics, thus belonging to the same group.
While there are several possible ways to define the pseudo-label, our approach more closely follows the ones based on argmax classification of the final layer \cite{kanezaki2018unsupervised,saha2019semantic}.
Once the pixels are assigned to the $K$ clusters, parameters of the deep network can be updated by using a loss between the 
feature $y^b_{1,n}$ and the cluster $c^b_{1,n}$. We use cross-entropy loss as:
\begin{equation}
\ell^b_{1,n}=\textrm{crossentropy}(y^b_{1,n},c^b_{1,n})
\end{equation}
In practice, the loss term $\mathcal{L}_1$ is computed by taking mean of $\ell^b_{1,n}$ over all pixels in $x^b_{1}$ and all patches in the batch ($b=1,...,\mathcal{B}$). 
$\mathcal{L}_1$ is used to adjust the weights of $h$ and $f_{opt}$. Similarly, 
$\mathcal{L}_2$ is computed from  $z^b_{2}$ ($b=1,...,\mathcal{B}$) and used to modulate the weights of $h$ and $f_{sar}$.
\par
While the deep clustering helps to learn representation for each sensor separately, they do not ensure that the independently learned features are aligned with each other.

\subsection{Temporal consistency}
\label{spatioTemporalHomogeneity}
Recalling from Section \ref{subsectionSiameseRepresentation}, multi-sensor bi-temporal patches $x^b_{1}$ and $z^b_{2}$ are 
multiple views of the same location in absence of any change. 
In other words, in  co-registered bi-temporal images, pixels in the same spatial location generally tend to belong to the same object as changes have
a low prior probability than the unchanged class. Thus, the features computed for the bi-temporal paired patches $x^b_{1}$ and $z^b_{2}$ should be 
similar in most cases. For each
input pixel $x^b_{1,n}$ and $z^b_{2,n}$,  we compute absolute error (AE) loss as:
\begin{equation}
\ell^b_{12,n}=||y^b_{1,n}-y^b_{2,n}||_{_1}
\end{equation}
A loss term $\mathcal{L}_{1,2}$ is computed by taking mean of $\ell^b_{12,n}$ over all considered pixels for all patches in the batch.
The proposed temporal consistency only ensures that the pixels at same location however at two different time tend to have same label. This may lead a to a degenerate solution
where all pixels simply have same prediction for both times. Moreover, some bi-temporal pairs $x^b_{1}$ and $z^b_{2}$ may be indeed changed, however, penalized for producing
dissimilar output in this step.

\begin{algorithm}[!t]
\caption{Self-supervised training for multisensor CD}\label{euclid}
\begin{algorithmic}[1]

\State Initialize $\mathbb{W}^1,...,\mathbb{W}^{L}$ 
\For{$i \gets 1$ to $\mathcal{I}$}    
\State Sample $\mathcal{B}$ patches from $X_{1}$, denoted as $\mathcal{X}=\{x^1_{1},...,x^\mathcal{B}_{1}\}$
\State Obtain corresponding $\mathcal{B}$ patches from $Z_{2}$, denoted as $\mathcal{Z}=\{z^1_{2},...,z^\mathcal{B}_{2}\}$ 
\State Obtain  $\mathcal{Z}'$ as random shuffling of $\mathcal{Z}$
	\For{$j \gets 1$ to $\mathcal{J}$}   
		\For {$b \in \mathcal{B}$}
			\State $y^b_1 = h(f_{opt}(x^b_{1}))$\\
			\State $y^b_2 = h(f_{sar}(z^b_{2}))$\\
			\State $y^{b'}_2 = h(f_{sar}(z^{b'}_{2}))$\\
		\EndFor
		\State Calculate deep clustering losses $\mathcal{L}_1$, $\mathcal{L}_2$ 
		\State Calculate temporal consistency loss $\mathcal{L}_{1,2}$
		\State Calculate contrastive loss $\mathcal{L}'_{1,2}$ 
		\If {$i \leq \mathcal{I}_1$}
			\State Use loss $(\mathcal{L}_1+\mathcal{L}_2)/2$ to modulate $\mathbb{W}^1,...,\mathbb{W}^{L}$
		\Else
			\State For each 3 consecutive iterations $j$, use  $\mathcal{L}_1$, $\mathcal{L}_{1,2}$, and $\mathcal{L}'_{1,2}$, respectively, to modulate $\mathbb{W}^1,...,\mathbb{W}^{L}$
		\EndIf		
	\EndFor
\EndFor
\end{algorithmic}
\label{algorithmProposedSelfSupLearningForMultisensorTraining}
\end{algorithm}

\subsection{Contrastive learning}
\label{spatioConstrastiveLoss}
While Section \ref{spatioTemporalHomogeneity} encourages the features computed for paired patch $x^b_{1}$ and $z^b_{2}$ to be similar, in this Section we encourage the network to produce dissimilar
feature for different input by employing concepts inspired from contrastive learning. While we do not have negative samples under the unsupervised setting in which our work is based on, we simply shuffle the 
batch of patches $\mathcal{Z}$ to $\mathcal{Z'}$. Recall that $\mathcal{X}$ and $\mathcal{Z}$ have location-wise paired patches. This implies that  $\mathcal{X}$ and $\mathcal{Z'}$ have unpaired patches.
Thus they should be more dissimilar in comparison to the paired patches in Section \ref{spatioTemporalHomogeneity}.
We encourage features computed for $x^b_{1}$ and $z^{b'}_{2}$ to be dissimilar. This is achieved by computing (negative) absolute error loss for each
input pixel $x^b_{1,n}$ and $z^{b'}_{2,n}$:
\begin{equation}
\ell^{b'}_{12,n}= -||y^b_{1,n}-y^{b'}_{2,n}||_{_1}
\end{equation}
$\ell^{b'}_{12,n}$ has negative value. Ideally $\ell^{b'}_{12,n}$ should be encouraged to be more and more negative. However in practice we note that, simply shuffling $\mathcal{Z}$ to $\mathcal{Z'}$ does
not always ensure that $\mathcal{X}$ and $\mathcal{Z'}$ have semantically different patches. Even after shuffling they may have the semantically paired patches, however penalized in this
step for producing similar features. Thus to control its impact, we penalize the network with $\ell^{b'}_{12,n}$ only when it approaches 0, i.e., $y^b_{1,n}$ and $y^{b'}_{2,n}$ become too similar. This
is achieved by computing the loss term $\mathcal{L}'_{1,2}$  as
mean of exponentials of $\ell^{b'}_{12,n}$ over all considered pixels for all patches in the batch.

\subsection{Overall loss and network refinement}
\label{sectionWeightRefinement}
He initialization process \cite{he2015delving} is used to initialize all the trainable weights of the network $\mathbb{W}^1,...,\mathbb{W}^{L}$, corresponding to $L$ layers. 
For updating of weights, we exploit stochastic gradient descent (SGD) mechanism  with momentum \cite{ruder2016overview}. The training process is executed in two different steps of $\mathcal{I}_1$ and
$\mathcal{I}_2$ epochs (summing to $\mathcal{I}$). For each batch of data, $\mathcal{J}$ iterations are performed.
For the first $\mathcal{I}_1$ epochs only sum of deep clustering losses ($\mathcal{L}_1+\mathcal{L}_2$) is used to modulate the network weights.
For subsequent $\mathcal{I}_2$ epochs, in one training iteration $\mathcal{L}_1$ is used as loss function, in the following iteration $\mathcal{L}_{1,2}$ is used and in the 
following iteration $\mathcal{L}'_{1,2}$ is used. The combination of three loss functions yield a balanced training process taking into account coherent cluster formation, temporal feature consistency, and
feature dissimilarity for unpaired patches. Alternatively, sum of  $\mathcal{L}_1$, $\mathcal{L}_{1,2}$, and $\mathcal{L}'_{1,2}$ can also be used as aggregated loss function.  The self-supervised
mechanism for network training is shown in Algorithm \ref{algorithmProposedSelfSupLearningForMultisensorTraining}.

\subsection{Change detection}
\label{sectionChangeDetection}
Once the network is trained, it can be used to detect change between $X_1$ and $Z_2$. Since the network is fully convolutional, it enables us to obtain pixelwise feature vector of dimension $K$
from $X_1$ and $Z_2$. Similar to \cite{saha2019unsupervised}, the pixelwise change information is captured by taking the magnitude
($\ell_2$ norm)  of difference of the feature vectors computed from pre-change and post-change pixels. Changed pixels ($\Omega_{c}$) generate higher difference magnitude in comparison to the unchanged ones
$\omega_{nc}$ and they can be distinguished by using any suitable threshold determination scheme \cite{otsu1979threshold}.

\begin{figure}[!t]
\centering
\subfigure[]{%
           \fbox{\includegraphics[height=2.5 cm]{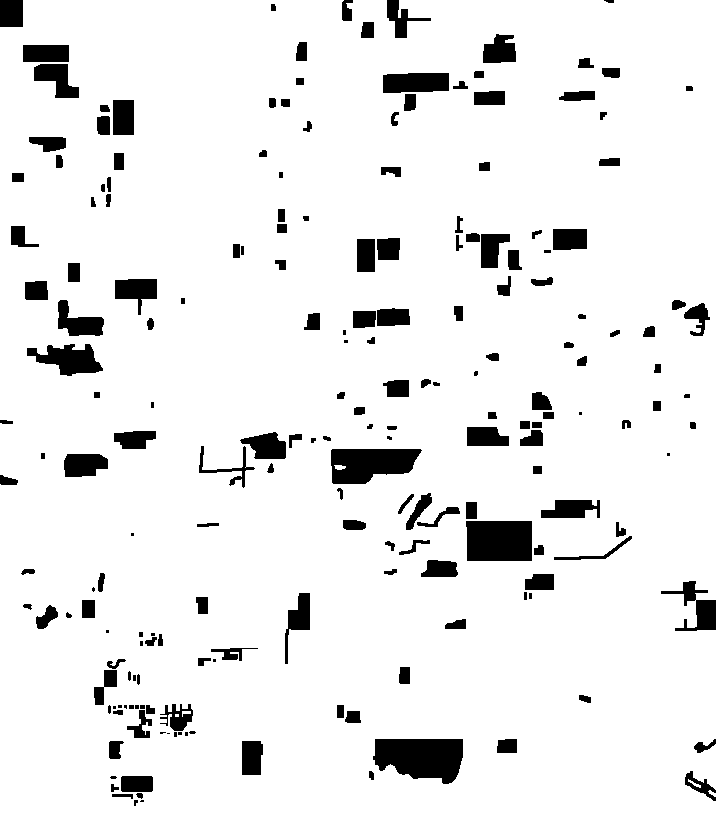}}
            \label{referenceImageLasvegas}
        }%
\hspace{0.2 cm}
\subfigure[]{%
           \fbox{\includegraphics[height=2.5 cm]{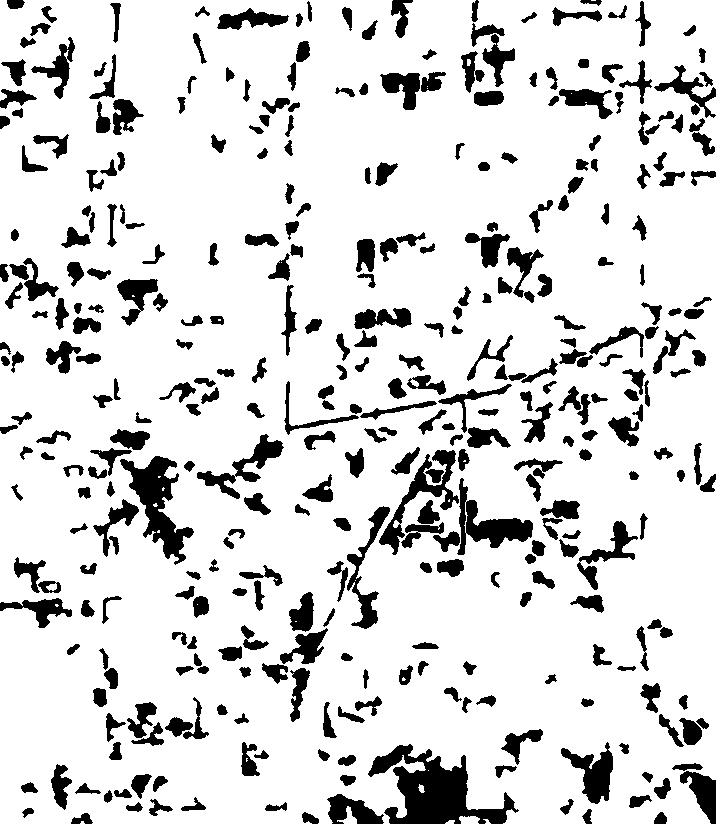}}
            \label{binaryChangeProposedMethodLasvegas}
        }%
\hspace{0.2 cm}
\subfigure[]{%
          \fbox{\includegraphics[height=2.5 cm]{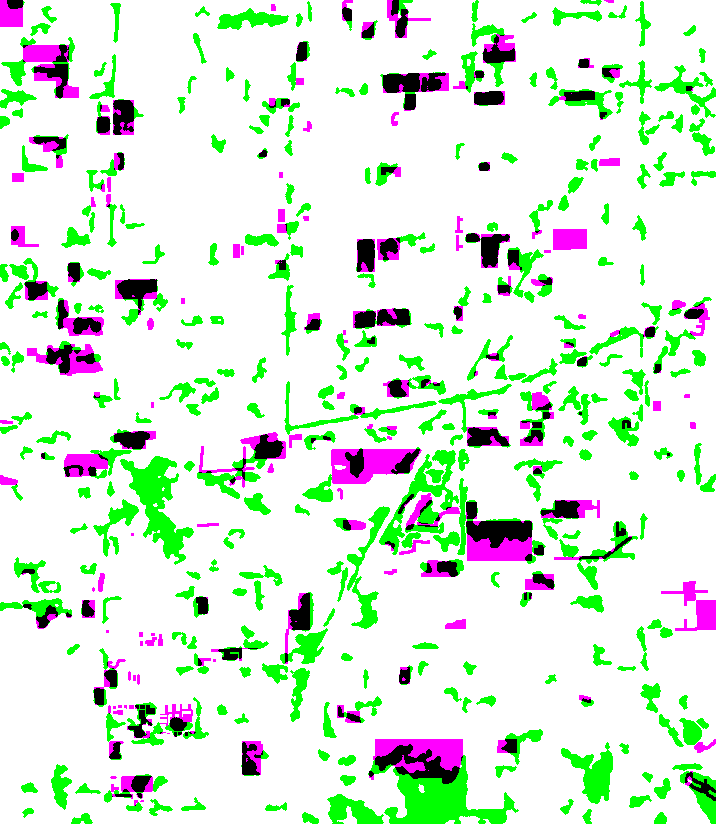}}
          \label{fccProposedMethodLasvegas}
       }%
\\
\subfigure[]{%
          \fbox{\includegraphics[height=2.5 cm]{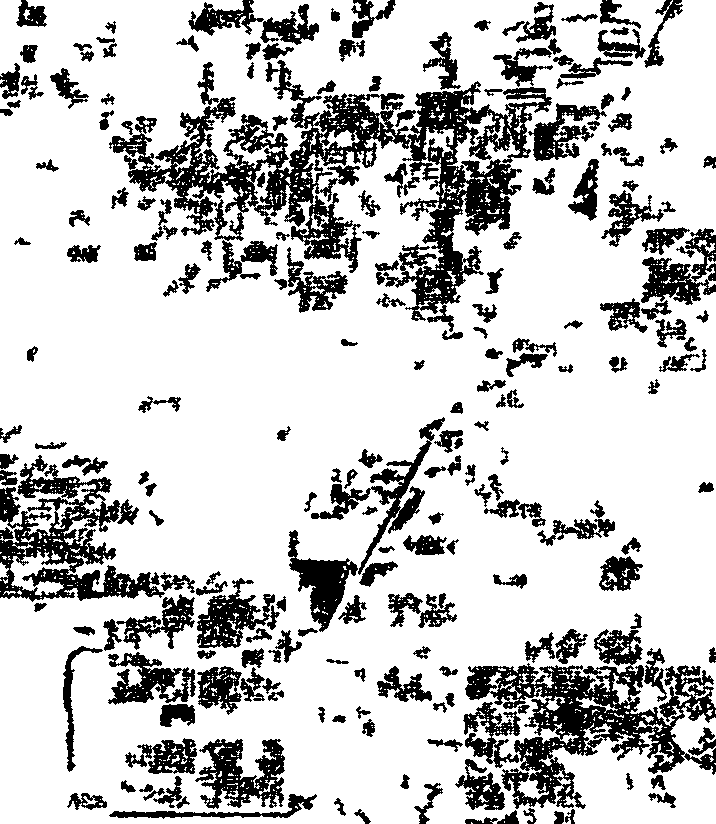}}
          \label{binaryChangeCvaLasvegas}
       }%
\hspace{0.2 cm}
\subfigure[]{%
          \fbox{\includegraphics[height=2.5 cm]{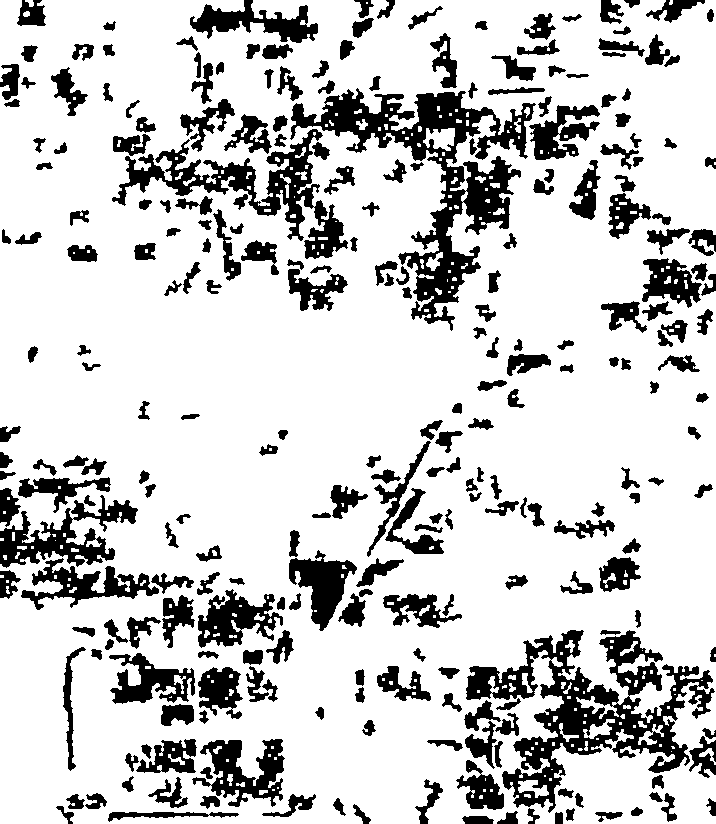}}
          \label{binaryChangeRcvaLasvegas}
       }%
\hspace{0.2 cm}
         \subfigure[]{%
          \fbox{\includegraphics[height=2.5 cm]{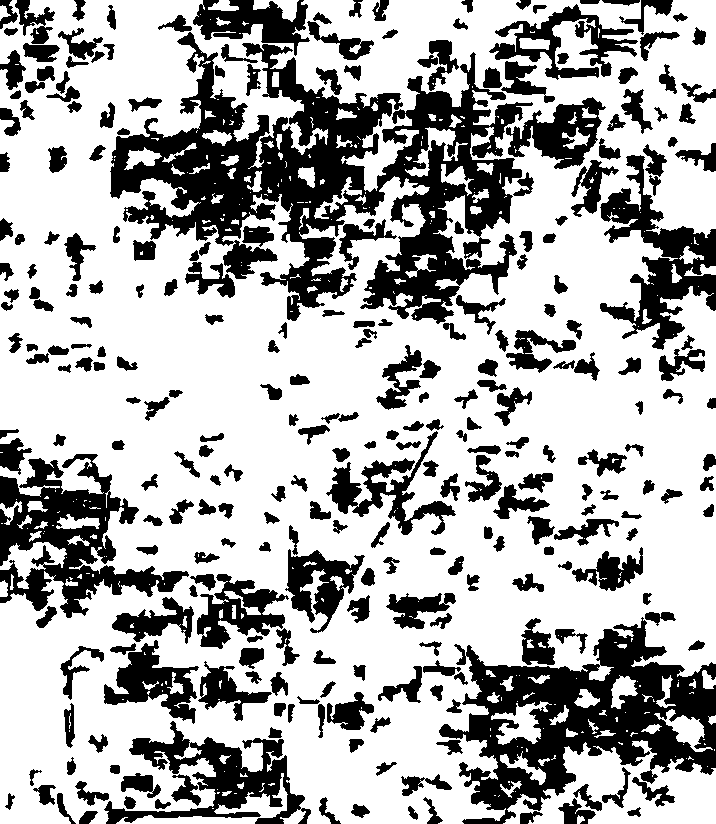}}
          \label{binaryChangePcvaLasvegas}
       }%
\\
\subfigure[]{%
          \fbox{\includegraphics[height= 2.5 cm]{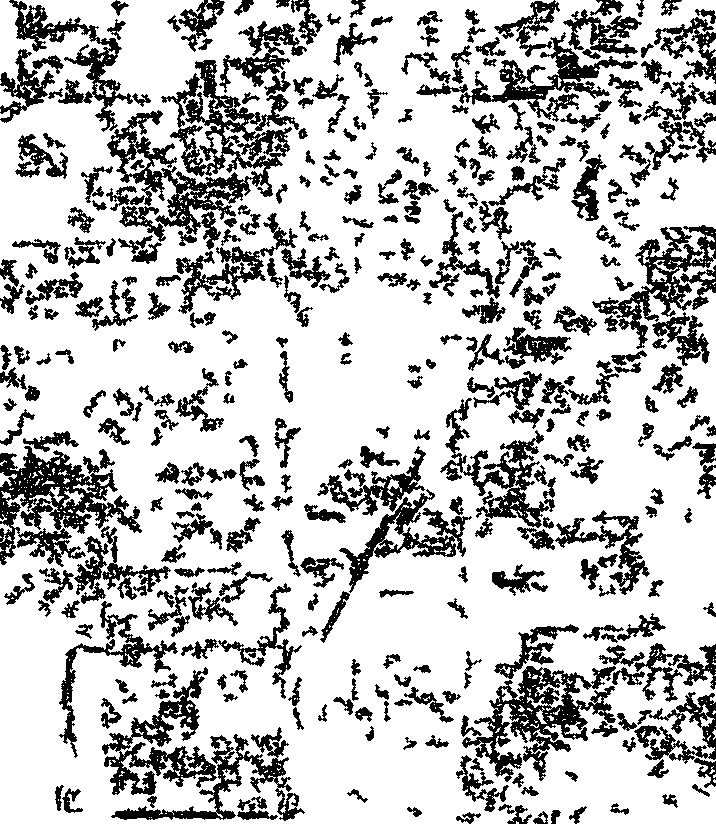}}
          \label{binaryChangeDcvaLasvegas}
       }%
        \hspace{0.2 cm}
\subfigure[]{%
          \fbox{\includegraphics[height=2.5 cm]{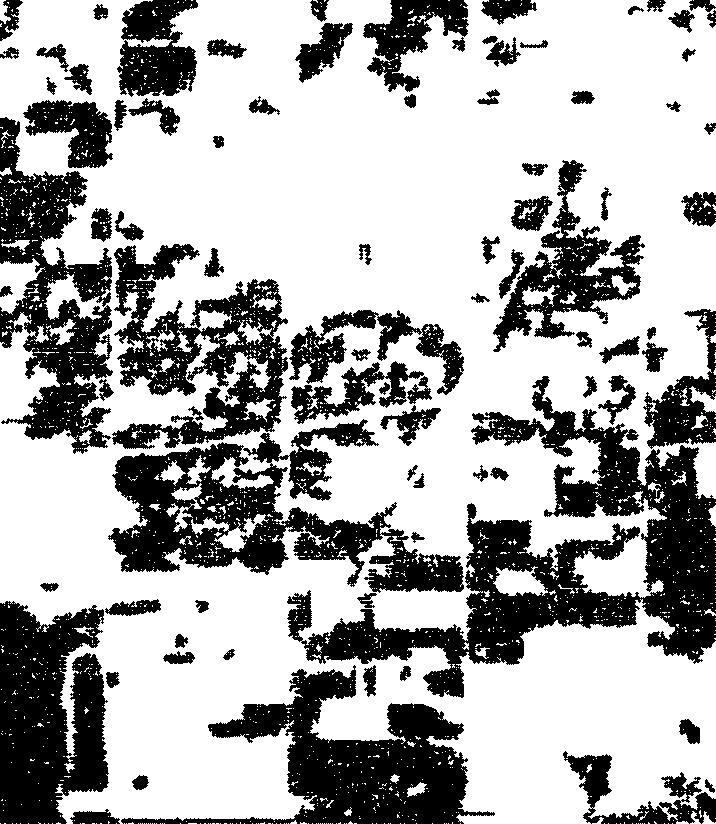}}
          \label{binaryChangeEncoderDecoderLasvegas}
       }%
\hspace{0.2 cm}
       \subfigure[]{%
          \fbox{\includegraphics[height=2.5 cm]{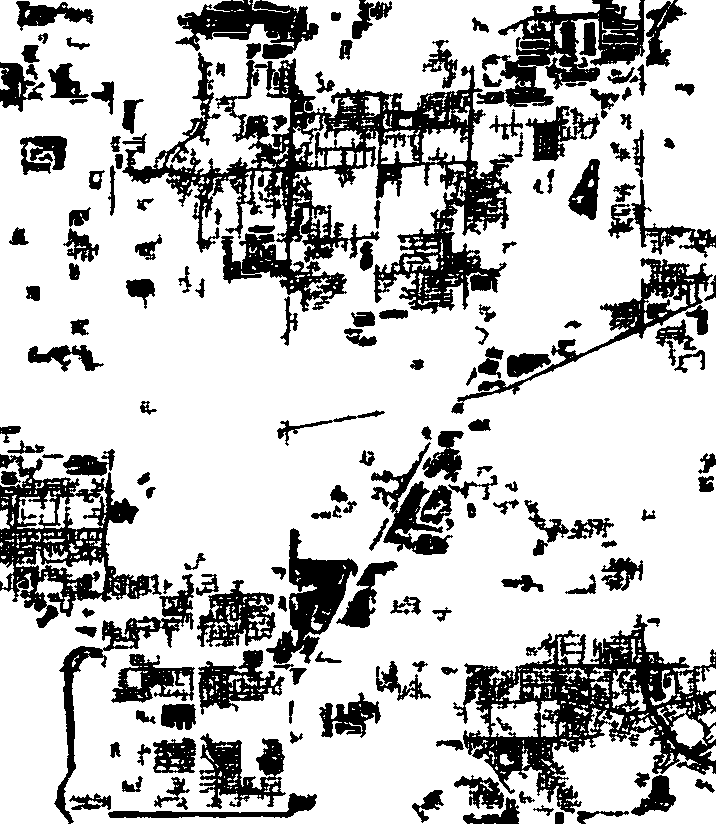}}
          \label{binaryChangeSccnLasvegas}
       }%
\vspace{-0.3cm}
\caption{CD results for Las Vegas. CD maps: (a) Reference, (b) Proposed, (c) FCC between reference and proposed (the correctly detected region are in black, false alarms are in green, missed alarms are in pink), (d) CVA, (e) RCVA, (f) PCVA,
(g) DCVA, (h) Encoder-decoder, and  (i) SCCN.}
\label{figureCdCityLasvegas}
\end{figure}

\begin{figure}[!t]
\centering
\subfigure[]{%
           \fbox{\includegraphics[height=2.5 cm]{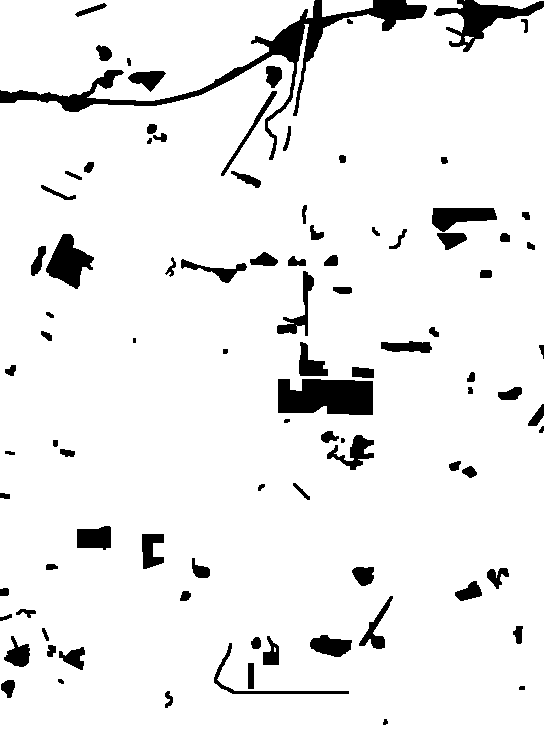}}
            \label{referenceImageChongqing}
        }%
\hspace{0.2 cm}
\subfigure[]{%
           \fbox{\includegraphics[height=2.5 cm]{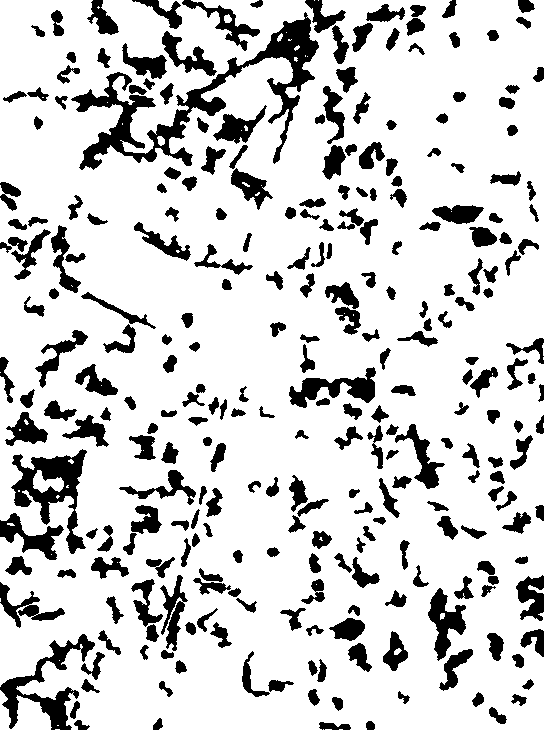}}
            \label{binaryChangeProposedMethodChongqing}
        }%
\hspace{0.2 cm}
\subfigure[]{%
          \fbox{\includegraphics[height=2.5 cm]{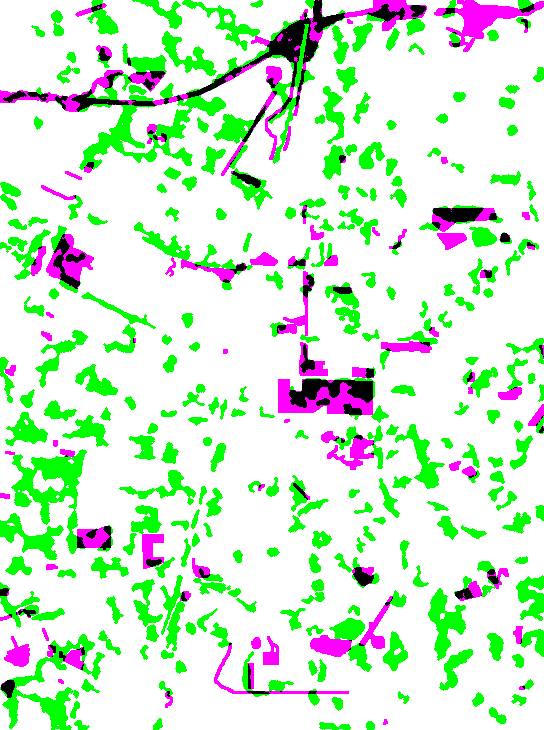}}
          \label{fccProposedMethodChongqing}
       }%
\\
\vspace{-0.3cm}
\subfigure[]{%
          \fbox{\includegraphics[height=2.5 cm]{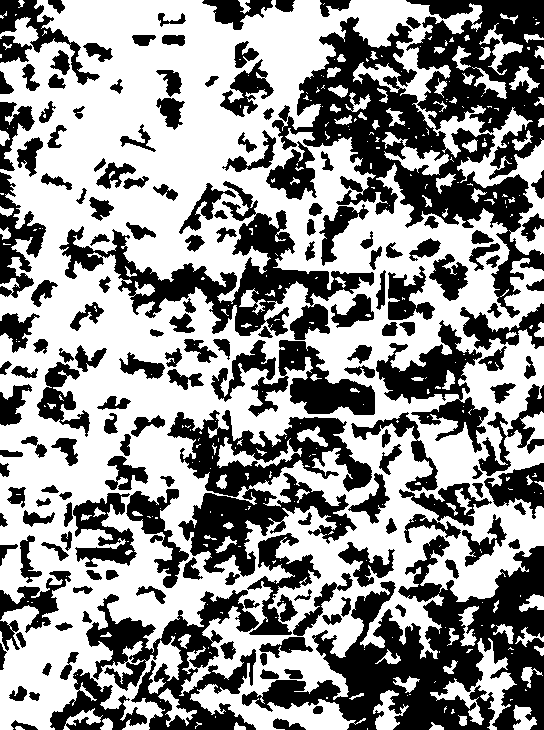}}
          \label{binaryChangePcvaChongqing}
       }%
\hspace{0.2 cm}
         \subfigure[]{%
          \fbox{\includegraphics[height=2.5 cm]{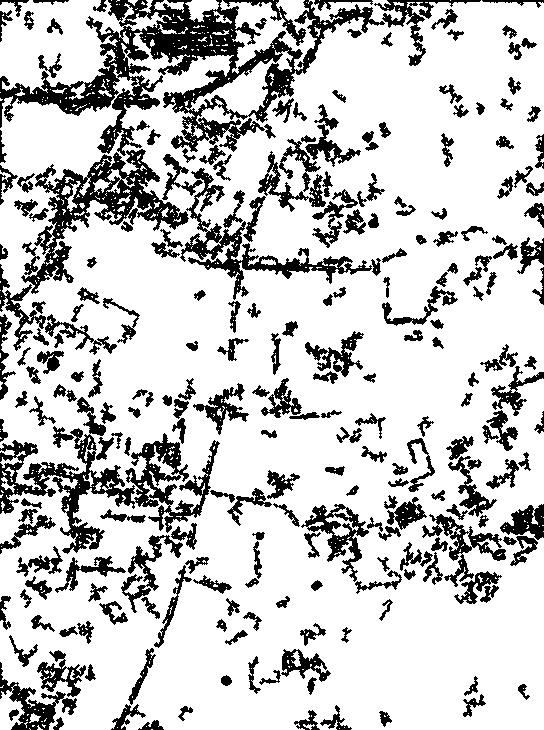}}
          \label{binaryChangeDcvaChongqing}
       }%
\hspace{0.2 cm}
\subfigure[]{%
          \fbox{\includegraphics[height=2.5 cm]{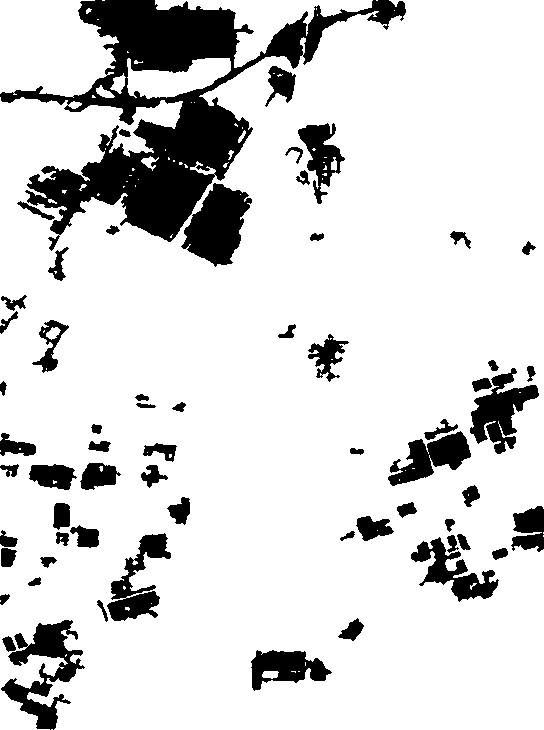}}
          \label{binaryChangeSccnChongqing}
       }%
\vspace{-0.3cm}
\caption{CD results for the Chongqing. CD maps: (a) Reference, (b) Proposed, (c) FCC between reference and proposed (the correctly detected region are in black, false alarms are in green, missed alarms are in pink), (d) PCVA, 
(e) DCVA, (f) SCCN.}
\label{figureCdCityChongqing}
\end{figure}

\begin{figure}[!t]
\centering
\subfigure[]{%
           \fbox{\includegraphics[height=2.5 cm]{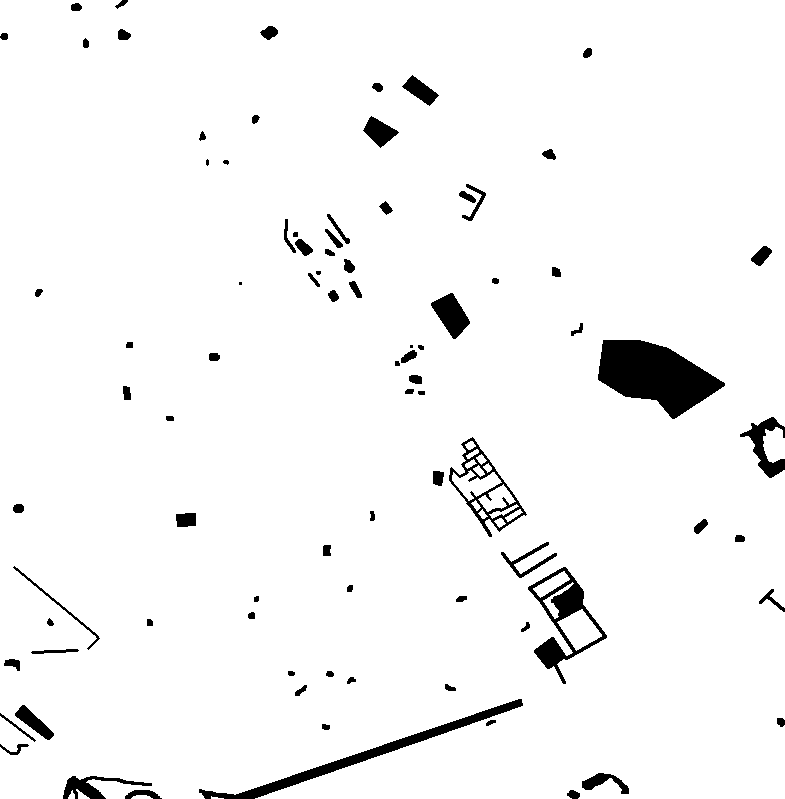}}
            \label{referenceImageAbudhabi}
        }%
\hspace{0.2 cm}
\subfigure[]{%
           \fbox{\includegraphics[height=2.5 cm]{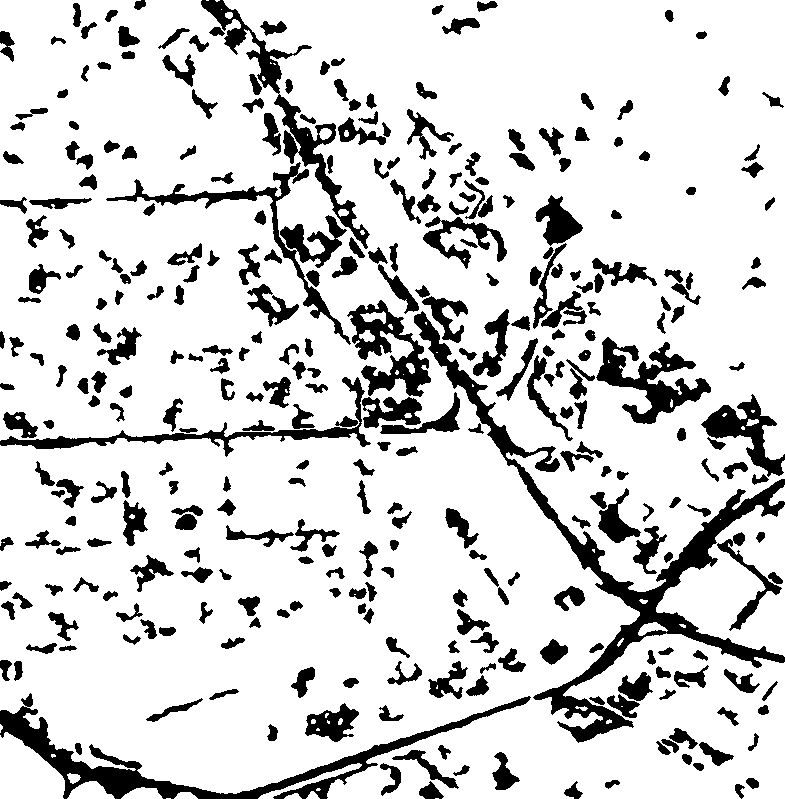}}
            \label{binaryChangeProposedMethodAbudhabi}
        }%
\hspace{0.2 cm}
\subfigure[]{%
          \fbox{\includegraphics[height=2.5 cm]{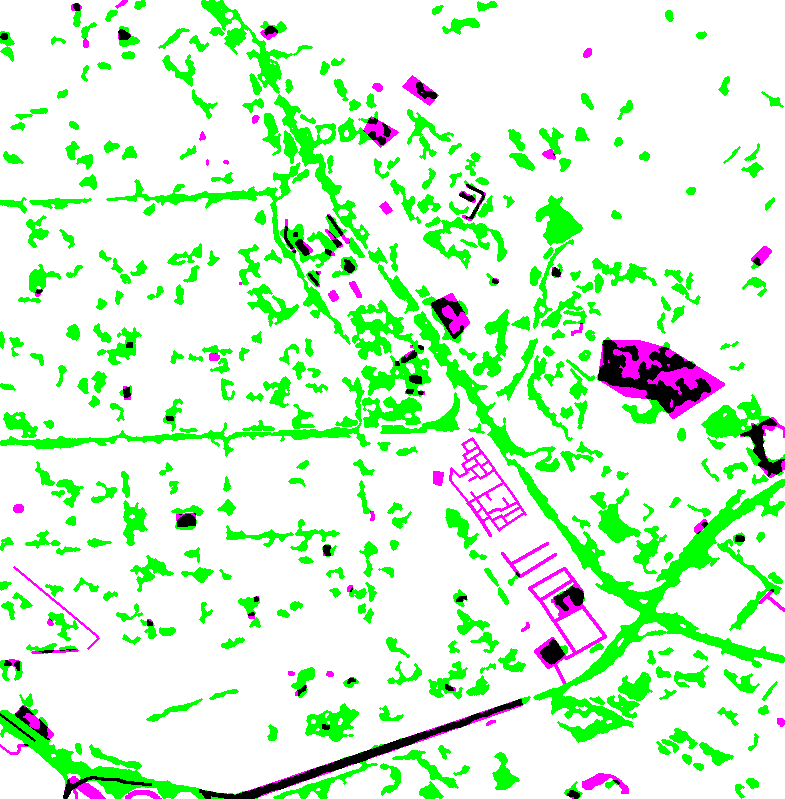}}
          \label{fccProposedMethodAbudhabi}
       }%
\\
\vspace{-0.3cm}
\subfigure[]{%
          \fbox{\includegraphics[height=2.5 cm]{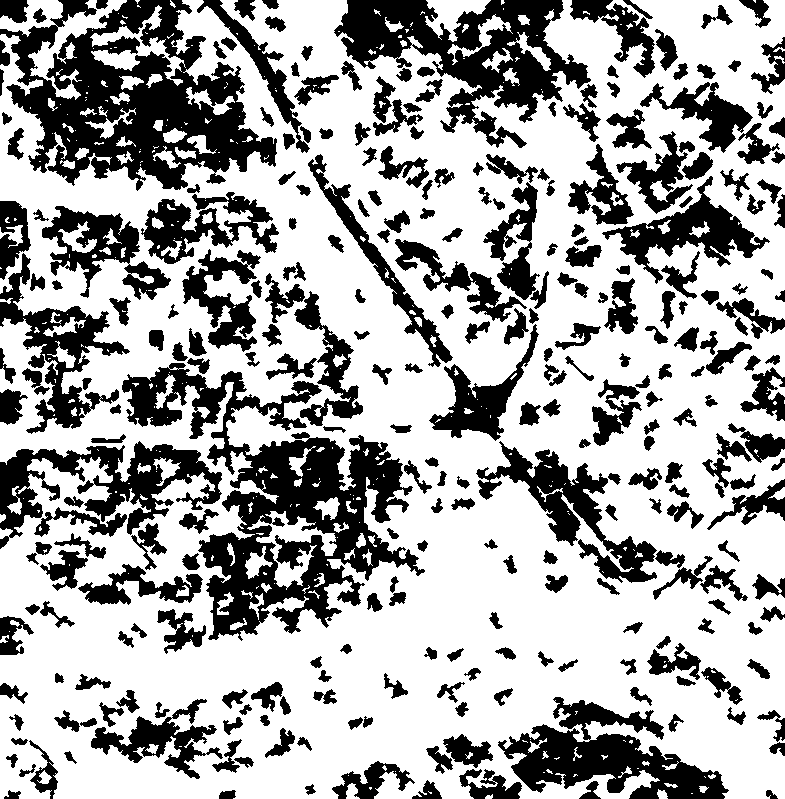}}
          \label{binaryChangePcvaAbudhabi}
       }%
\hspace{0.2 cm}
         \subfigure[]{%
          \fbox{\includegraphics[height=2.5 cm]{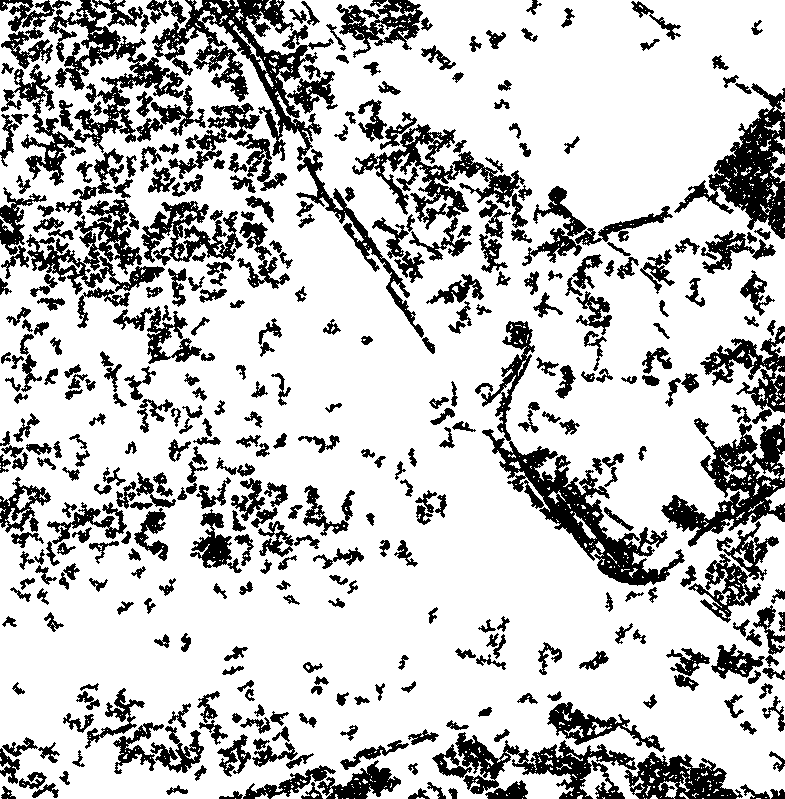}}
          \label{binaryChangeDcvaAbudhabi}
       }%
\hspace{0.2 cm}
\subfigure[]{%
          \fbox{\includegraphics[height=2.5 cm]{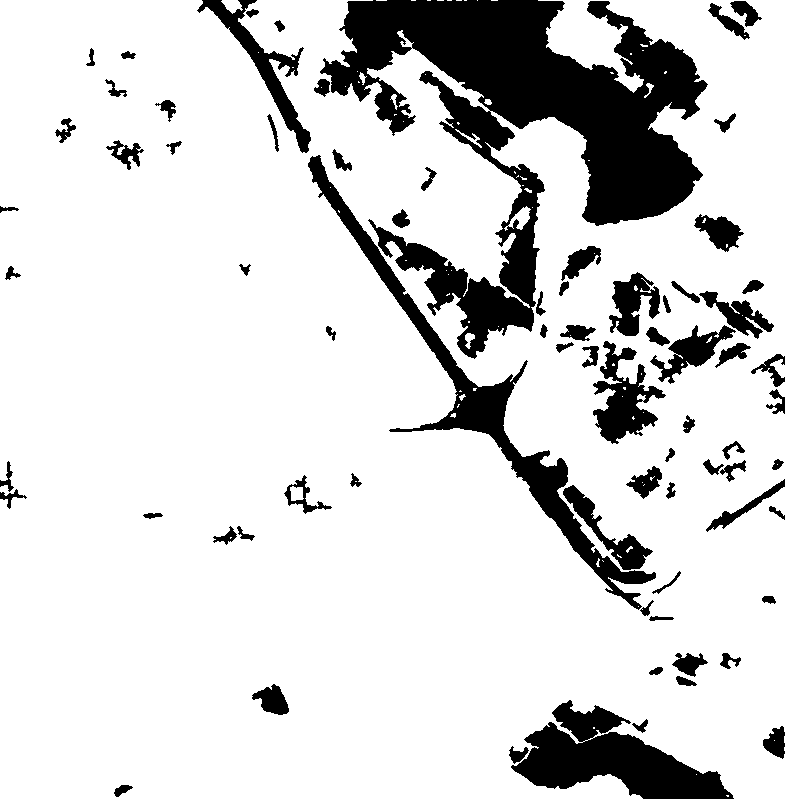}}
          \label{binaryChangeSccnAbudhabi}
       }%
\vspace{-0.3cm}
\caption{CD results for the Abu Dhabi. CD maps: (a) Reference, (b) Proposed, (c) FCC between reference and proposed (the correctly detected region are in black, false alarms are in green, missed alarms are in pink), (d) PCVA, 
(e) DCVA, (f) SCCN.}
\label{figureCdCityAbudhabi}
\end{figure}

\begin{figure}[!ht]
\centering
\subfigure[]{%
           \fbox{\includegraphics[height=2.1 cm]{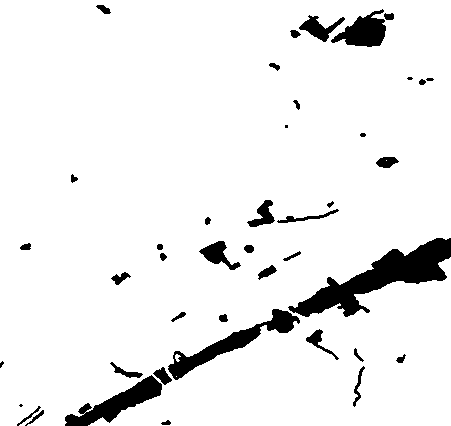}}
            \label{referenceImageMontpellier}
        }%
\hspace{0.2 cm}
\subfigure[]{%
           \fbox{\includegraphics[height=2.1 cm]{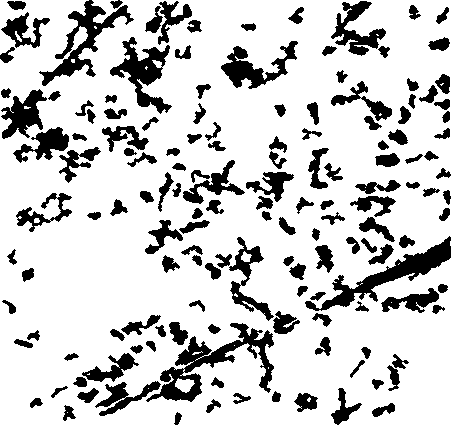}}
            \label{binaryChangeProposedMethodMontpellier}
        }%
\hspace{0.2 cm}
\subfigure[]{%
          \fbox{\includegraphics[height=2.1 cm]{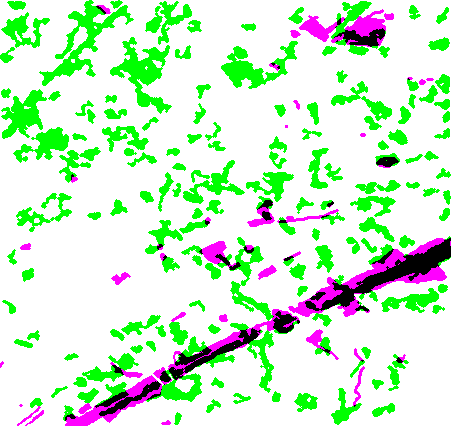}}
          \label{fccProposedMethodMontpellier}
       }%
\\
\vspace{-0.3cm}
\subfigure[]{%
          \fbox{\includegraphics[height=2.1 cm]{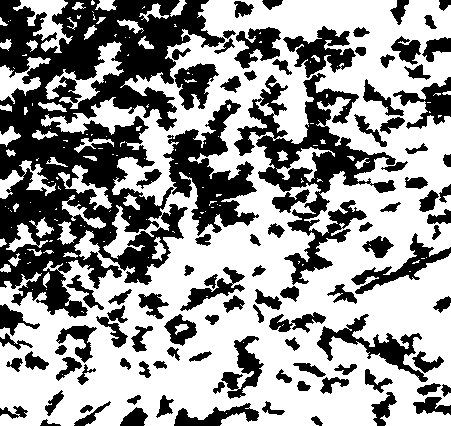}}
          \label{binaryChangePcvaMontpellier}
       }%
\hspace{0.2 cm}
         \subfigure[]{%
          \fbox{\includegraphics[height=2.1 cm]{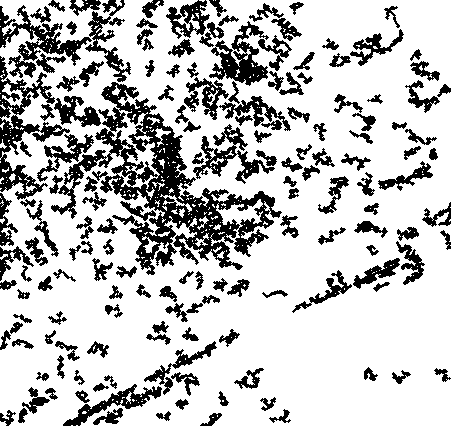}}
          \label{binaryChangeDcvaMontpellier}
       }%
\hspace{0.2 cm}
\subfigure[]{%
          \fbox{\includegraphics[height=2.1 cm]{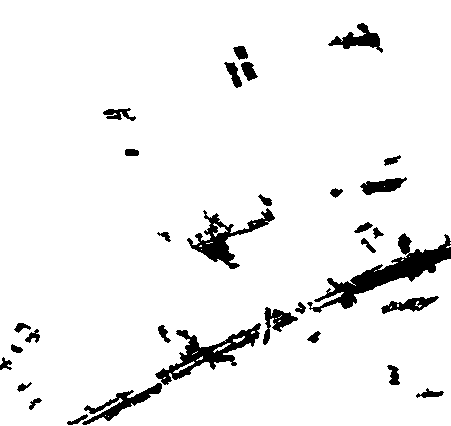}}
          \label{binaryChangeSccnMontpellier}
       }%

\vspace{-0.3cm}
\caption{Qualitative CD results for the Montpellier. Change detection maps: (a) Reference, (b) Proposed, (c) FCC between reference and proposed (the correctly detected region are in black, false alarms are in green, missed alarms are in pink), (d) PCVA, 
(e) DCVA, (f) SCCN.}
\label{figureCdCityMontpellier}
\end{figure}

\section{Experimental Validation}
\label{sectionExperimentalResult}
\subsection{Datasets}
We use four paired optical (pre-change) - SAR (post-change) images to validate the proposed method. Optical images are acquired by the Sentinel-2 sensor and are taken from the
Onera Satellite Change Detection (OSCD) dataset \cite{daudt2018urban}. They show 10 m/pixel spatial resolution. OSCD dataset
is originally a single-sensor dataset consisting only Sentinel-2 images. Recalling the importance of multisensor CD (see Section \ref{secIntro}), we extend this dataset by collecting the 
post-change SAR Sentinel-1 images for the nearest available date as the post-change image in original OSCD dataset. 
Both Sentinel-2 and Sentinel-1 sensors are part of the European Space Agency's Copernicus program.
\par
The four scenes are collected over Las Vegas in United States (824 $\times$ 716 pixels) (Figure \ref{figureCdCityLasvegas}), Chongqing in China (730 $\times$ 544 pixels)
(Figure \ref{figureCdCityChongqing}), Abu Dhabi (799 $\times$ 785 pixels) (Figure \ref{figureCdCityAbudhabi}), and Montpellier in France (426 $\times$ 451 pixels) (Figure \ref{figureCdCityMontpellier}). Thus
this provides us an opportunity to validate the proposed method on geographically distributed 
complex urban scenes with large variation.

\subsection{Compared methods}
To verify the effectiveness of the proposed method, we compare it to related unsupervised change detection methods:
\begin{enumerate}
\item Change vector analyis (CVA) \cite{malila1980change, bruzzone2000automatic}, a classical difference-based unsupervised model for change detection.
\item Robust change vector analysis (RCVA) \cite{thonfeld2016robust} that modifies CVA by taking into account pixel neighborhood effects.
\item Parcel change vector analysis (PCVA) \cite{bovolo2009multilevel} that incoporates notion of object (superpixels) in CVA.
\item Deep change vector analysis (DCVA) \cite{saha2019unsupervised} that detects change by comparing bi-temporal deep features extracted using a pre-trained network.  We used second
convolution layer of pre-trained VGGNet \cite{simonyan2014very} for feature extraction.
\item Image-to-image transfer model based on an encoder-decoder network architecture that projects pre-change optical images into post-change SAR image \cite{xu2013change}. CD map
can be obtained by difference of the simulated pre-change SAR image (obtained as projection of
pre-change optical image) and the original post-change SAR image. 
\item Denoising autoencoder (DAE) based joint feature extraction \cite{zhan2018log}.
\item SCCN  \cite{liu2016deep} that first identifies some unchanged pixels and uses them to learn a coupled network.
\end{enumerate}
While the methods 1-3 are not deep learning based, the following ones are deep learning based. The methods 1-4 do not have any explicit adaptation for multi-sensor input, while the method 5, 6, and 7 have.

\subsection{Experimental settings}
The proposed method and compared methods are fed with pre-processed images and post-processed
similarly. For the proposed method, we use $\mathcal{I} = 5$ ($\mathcal{I}_1 = 1$, $\mathcal{I}_2 = 4$), $\mathcal{J} = 50$ $K =4$, $L_1 =4$, and $L_2 =1$. 
We show  the architecture of the network in Table \ref{theNetworkOneBranchStructure}. A relatively simple architecture is used considering that the number of patches available
to us is very few compared to the images in typical computer vision datasets. Moreover, our target image has coarse resolution (10 m/pixel) compared to natural images in
computer vision. Spatial complexity in such coarse images can be handled by simpler architecture compared to those in computer vision. 
64 $\times$ 64 patches are used to train the model and patches
are extracted from the bitemporal scene with a stride of 32. The actual number of training patches for a 
scene depends on the size of the particular scene. E.g., for the Las Vegas scene (824 $\times$ 716 pixels), number of 
patches extracted is 504. For optimization, Stochastic Gradient Descent method is used with learning rate set to 0.001.

\begin{table}[h]
\center
\caption{Structure of the network for processing one of the two inputs}
\begin{tabular}{|c|c|c|c|} 
\hline
\textbf{Layer} & \textbf{Kernel number} & \textbf{Kernel size} & \textbf{Stride}\\ 
\hline
convolution & 64 & (3,3) & 1\\ 
\hline
convolution & 64 & (3,3) & 1\\ 
\hline
convolution & 64 & (3,3) & 1\\
\hline
convolution & 64 & (3,3) & 1\\
\hline
convolution & $K$ & (1,1) & 1 \\
\hline
\end{tabular}
\label{theNetworkOneBranchStructure}
\end{table}

\par
We show result in terms of
sensitivity (accuracy in percentage computed over reference changed pixels) and specificity (computed
over reference unchanged pixels). In more details, given true positive (TP), true negative (TN), false positive (FP) and false negative (FN), sensitivity is TP$/$(TP+FN) and
specificity is TN$/$(TN+FP).

\subsection{Results}

\textbf{Las Vegas:} The reference CD map (ground truth) for Las Vegas is shown in Figure \ref{referenceImageLasvegas}. Figure \ref{binaryChangeProposedMethodLasvegas}
shows the result obtained by the proposed method. For better visualization, a false color composition between the reference map and the obtained result is shown
in Figure \ref{fccProposedMethodLasvegas}. The proposed method can detect most of the changed objects with fewer false alarms in comparison to the
compared methods. In many cases, proposed method partly detects the changed object, thus missing some objects only partially (shown in pink in Figure \ref{fccProposedMethodLasvegas}).
CVA (Figure \ref{binaryChangeCvaLasvegas}) performs poorly and incorrectly detects most urban area as changed. Result obtained by RCVA (Figure \ref{binaryChangeRcvaLasvegas}) is similar to CVA. While
PCVA (Figure \ref{binaryChangePcvaLasvegas}), DCVA (Figure \ref{binaryChangeDcvaLasvegas}), encoder-decoder (Figure \ref{binaryChangeEncoderDecoderLasvegas}), DAE, and SCCN (Figure \ref{binaryChangeSccnLasvegas}) improve
 the result over CVA, 
proposed method still outperforms them by large margin.  Quantitative evaluation (Table \ref{tableLasVegasComparison}) clearly shows the superiority of the proposed method over state-of-the-art
 unsupervised methods. This can be attributed to superior capability of the proposed method to ingest
multi-sensor multi-temporal images.
\par
Further studies  are conducted by varying different parameters on Las Vegas image pair.
\par
\textit{Training epochs}  $\mathcal{I}$ is varied with different values as tabulated in Table \ref{tableLasVegasVariationWithIter} while setting
$K = 4 $. 
We observe clear improvement in performance from $\mathcal{I} = 1$ to 2. Recalling from Section \ref{sectionWeightRefinement} that for first $\mathcal{I}_1 = 1$ iterations  only deep clustering loss is used, this shows that
bi-temporal deep clustering itself is not sufficient to learn the correspondence between two images and the other losses ($\mathcal{L}_{1,2}$ and $\mathcal{L}'_{1,2}$) are required.
$\mathcal{I} = 2$ onwards, we observe an increment in performance initially followed by performance getting saturated/dropping. Despite variation in performance, proposed method outperforms all compared methods for
 $\mathcal{I} = 3, 5, 10$.
\par
\textit{Kernel number of last layer} ($K$) is varied from 2 to 16 in multiplicative steps of 2 while fixing
the $\mathcal{I} = 5 $. The variation in performance is shown in Table \ref{tableLasVegasVariationWithK}. While performance improves from $K = 2$ to $K = 4$,  a 
gradual fall in performance is observed henceforth. Increasing value of $K$ is equivalent to allowing the scene to be partitioned more classes. Since the spatial area of the scene is fixed and
not too large (only few hundred pixels by few hundred pixels), large number of classes potentially leads the model to learn irrelevant classes, impacting change detection performance.
\par
\textit{Thresholding} is done using Otsu's method \cite{otsu1979threshold}, as it is popular in unsupervised CD methods \cite{thonfeld2016robust,saha2020unsupervisedGrsl}.  However any other suitable method can be used, as shown in Table \ref{tableLasVegasVariationThresholdScheme}. Results obtained by ISODATA method \cite{ridler1978picture,sezgin2004survey} and adaptive method \cite{saha2019unsupervised} are similar to Otsu's method \cite{otsu1979threshold}. 
\par
\begin{figure}[t]
\centering
\subfigure[]{%
           \includegraphics[height=2.6 cm]{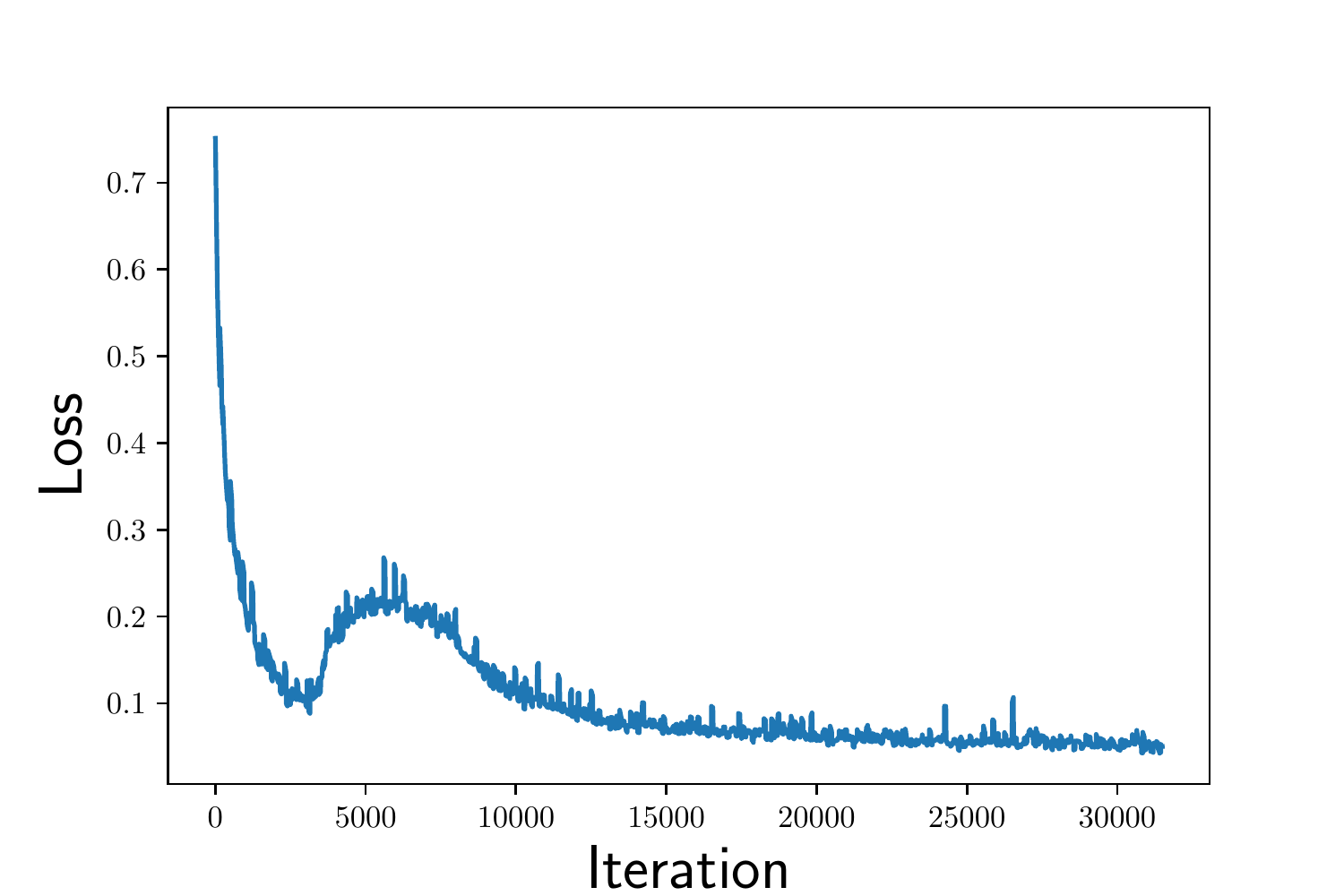}
            \label{primaryLossPlotLasvegas}
        }%
\hspace{-0.1cm}
\subfigure[]{%
            \includegraphics[height=2.6 cm]{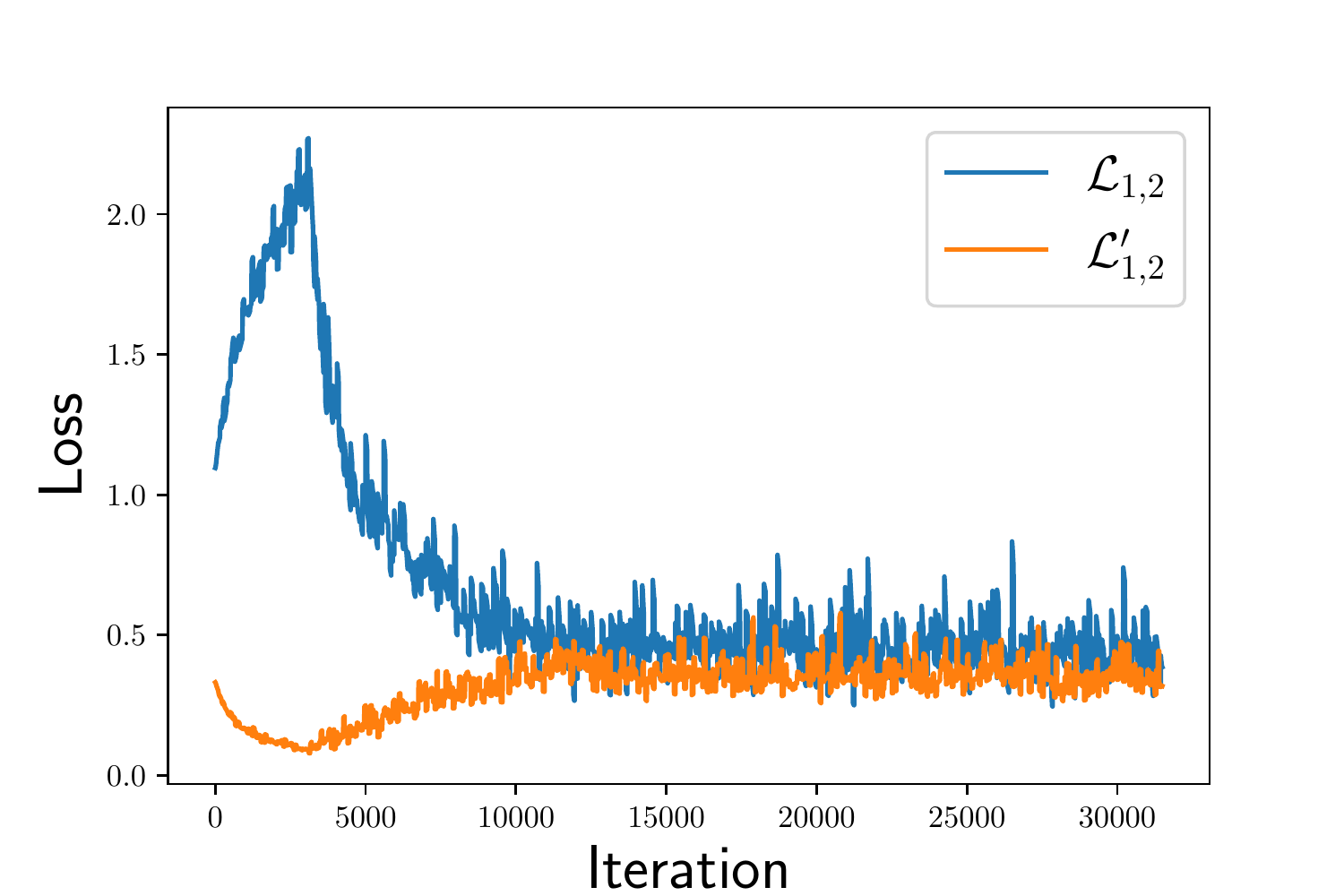}
            \label{secondaryLossPlotLasvegas}
        }%
\caption{Evolution of the loss over training iterations for Las Vegas: (a) deep clustering loss $\mathcal{L}_{1}$, (b) temporal consistency loss $\mathcal{L}_{1,2}$ and contrastive loss $\mathcal{L}'_{1,2}$}
\label{figureLossVisualizationLasvegas}
\end{figure}

\textit{Loss plot visualization} in Figure \ref{figureLossVisualizationLasvegas} shows the interplay between different components of loss. 
$\mathcal{L}_1$ consistently decreases (Figure \ref{primaryLossPlotLasvegas}) except it rises for a while after epoch 1 when $\mathcal{L}_{1,2}$ and $\mathcal{L}'_{1,2}$ are introduced to the training process.
$\mathcal{L}_{1,2}$ and $\mathcal{L}'_{1,2}$ balances each other as shown in Figure \ref{secondaryLossPlotLasvegas}.
\par
\textit{Projection layers} $f_{opt}$ and $f_{sar}$ need to modeled
independently by not sharing weights between them to capture the different semantic properties of
optical and SAR patches, as hypothesized in Section \ref{subsectionSiameseRepresentation}.  Here we test this hypothesis by instead sharing
the weights between $f_{opt}$ and $f_{sar}$. For $\mathcal{I} = 5$
and $K =4$ the proposed method fails
to detect most of the changes. This shows that it is crucial to model the optical and SAR patches differently.
\par
\textit{Computation time} requirement is not high. We tested our code on a machine equipped with a Quadro T2000 GPU, which is a low end GPU. For processing Las Vegas dataset (training process over 5 epochs), it takes
approx. 460 seconds. Las Vegas scene is 824 $\times$ 716 pixels with 10 m/pixel resolution and thus processing it is equivalent to processing an approximate area of 8*7 = 56 sq. km in terms of geography.
\par
\textit{Same sensor} bi-temporal input can be ingested by the proposed method, though designed for multi-sensor CD. For Las Vegas prechange optical - postchange optical input, proposed method can obtain a sensitivity of
$64.74\%$ and specificity of $97.89\%$. However, we note that some characteristics of the proposed method (e.g., temporal consistency loss) are designed to reduce the representation gap of multi-sensor input, which
is less relevant in single-sensor input. So the proposed method may not be the most suitable choice for single-sensor scenarios as there are numerous existing CD techniques particularly
designed for same-sensor scenario \cite{saha2019unsupervised}.

\textbf{Chongqing and Abu Dhabi:} Reference CD map (ground truth) for Chongqing is shown in Figure \ref{referenceImageChongqing}. Figure \ref{binaryChangeProposedMethodChongqing} and \ref{fccProposedMethodChongqing}
show the result obtained by the proposed method and  false color composition between the reference map and the obtained result, respectively.  Proposed method 
outperforms all compared methods, as can be observed in quantitative result in Table \ref{tableChongqingComparison}). Similar result is obtained
for Abu Dhabi (Figure \ref{figureCdCityAbudhabi} and Table \ref{tableAbudhabiComparison}).
 \par
\textbf{Montpellier:} Reference CD map (ground truth) for Montpellier is shown in Figure \ref{referenceImageMontpellier}. The proposed method (Figure \ref{binaryChangeProposedMethodMontpellier})  outperforms most of the
 state-of-the-art methods including PCVA (Figure \ref{binaryChangePcvaMontpellier}) and DCVA (Figure \ref{binaryChangeDcvaMontpellier}), as shown
  in Table \ref{tableMontpellierComparison}. However, SCCN (Figure \ref{binaryChangeSccnMontpellier}) outperforms the proposed method.
The performance of the proposed method is relatively poor for Montpellier, which can be possibly explained by: 1) smaller size of Montpellier scene, which implies 
less data to learn proposed self-supervised network and 2) uniform (showing mostly urban areas) geospatial characteristics of Montpellier scene in comparison 
to Las Vegas and Chongqing that
show complex distribution consisting of both urban and non-urban areas.
 \par

\renewcommand{\tabcolsep}{2pt}
\begin{table}
\centering
\caption{Comparison of different methods on Las Vegas}
\begin{tabular}{|c|c|c|} 
 \hline
\textbf{Method} & \textbf{Sensitivity}  & \textbf{Specificity}  \\ 
\hline
\bf Proposed & 50.28 & 88.06 \\ 
\hline
\bf CVA & 9.64 & 77.13  \\ 
\hline
\bf RCVA  & 8.65 & 78.97  \\
\hline
\bf PCVA  & 23.60 & 67.92  \\
\hline
\bf DCVA  & 20.67 & 75.40  \\
\hline
\bf Encoder-decoder  & 46.30 & 68.58  \\
\hline
\bf DAE  & 39.07 & 83.72  \\
\hline
\bf SCCN  & 24.58 & 75.30  \\
\hline
  \end{tabular}
\label{tableLasVegasComparison}
\end{table}

\renewcommand{\tabcolsep}{2pt}
\begin{table}
\centering
\caption{Variation of result for Las Vegas as $\mathcal{I}$ is varied}
\begin{tabular}{|c|c|c|} 
 \hline
\textbf{$\mathcal{I}$} & \textbf{Sensitivity}  & \textbf{Specificity}  \\ 
\hline
\bf 0 (Just initialized) & 32.21 & 85.30 \\ 
\hline
\bf 1 & 61.70 & 53.40  \\ 
\hline
\bf 2  & 52.20 & 79.82  \\
\hline
\bf 3  & 46.30 & 86.52  \\
\hline
\bf 5  & 50.28 & 88.06  \\
\hline
\bf 10  & 41.79 & 88.80  \\
\hline
  \end{tabular}
\label{tableLasVegasVariationWithIter}
\end{table}

\renewcommand{\tabcolsep}{2pt}
\begin{table}
\centering
\caption{Variation of result for Las Vegas as $K$ is varied}
\begin{tabular}{|c|c|c|} 
 \hline
$K$ & \textbf{Sensitivity}  & \textbf{Specificity}  \\ 
\hline
\bf 2 & 41.47 & 87.48 \\ 
\hline
\bf 4 & 50.28 & 88.06  \\ 
\hline
\bf 8  & 42.75 & 84.70  \\
\hline
\bf 16  & 60.01 & 77.86  \\
\hline
  \end{tabular}
\label{tableLasVegasVariationWithK}
\end{table}

\renewcommand{\tabcolsep}{2pt}
\begin{table}
\centering
\caption{Variation of result for Las Vegas as threshold determination scheme is varied}
\begin{tabular}{|c|c|c|} 
 \hline
\textbf{Thresholding} & \textbf{Sensitivity}  & \textbf{Specificity}  \\ 
\hline
\bf Otsu & 50.28 & 88.06  \\ 
\hline
\bf ISODATA  & 50.48 & 87.95  \\
\hline
\bf Adaptive  & 50.19 & 83.65  \\
\hline
  \end{tabular}
\label{tableLasVegasVariationThresholdScheme}
\end{table}

\renewcommand{\tabcolsep}{2pt}
\begin{table}
\centering
\caption{Comparison of different methods on Chongqing}
\begin{tabular}{|c|c|c|} 
 \hline
\textbf{Method} & \textbf{Sensitivity}  & \textbf{Specificity}  \\ 
\hline
\bf Proposed & 36.17 & 83.26 \\ 
\hline
\bf CVA & 40.70 & 48.82  \\ 
\hline
\bf RCVA  & 41.96 & 44.28  \\
\hline
\bf PCVA  & 35.15 & 56.76  \\
\hline
\bf DCVA  & 32.67 & 79.18  \\
\hline
\bf Encoder-decoder  & 19.59 & 82.23  \\
\hline
\bf DAE  & 17.58 & 82.60  \\
\hline
\bf SCCN  & 30.67 & 85.73 \\
\hline
  \end{tabular}
\label{tableChongqingComparison}
\end{table}

\renewcommand{\tabcolsep}{2pt}
\begin{table}
\centering
\caption{Comparison of different methods on Abu Dhabi}
\begin{tabular}{|c|c|c|c|} 
 \hline
\textbf{Method} & \textbf{Sensitivity}  & \textbf{Specificity}  \\ 
\hline
\bf Proposed & 48.92 & 84.38 \\ 
\hline
\bf CVA  & 4.18 & 73.09  \\
\hline
\bf RCVA  & 5.75 & 73.47  \\
\hline
\bf PCVA  & 13.30 & 65.24  \\
\hline
\bf DCVA  & 20.35 & 76.74 \\
\hline
\bf Encoder-decoder  & 36.52 & 74.22  \\
\hline
\bf DAE  & 46.29 & 72.29  \\
\hline
\bf SCCN  & 9.79 & 83.88  \\
\hline
  \end{tabular}
  \label{tableAbudhabiComparison}
\end{table}

\renewcommand{\tabcolsep}{2pt}
\begin{table}
\centering
\caption{Comparison of different methods on Montpellier}
\begin{tabular}{|c|c|c|} 
 \hline
\textbf{Method} & \textbf{Sensitivity}  & \textbf{Specificity}  \\ 
\hline
\bf Proposed & 43.05 & 81.88 \\ 
\hline
\bf CVA & 9.09 & 74.60 \\ 
\hline
\bf RCVA  & 8.76 & 72.31  \\
\hline
\bf PCVA  & 25.74 & 58.59  \\
\hline
\bf DCVA  & 32.37 & 76.49  \\
\hline
\bf Encoder-decoder  & 46.10 & 75.52  \\
\hline
\bf DAE  & 24.46 & 73.97  \\
\hline
\bf SCCN  & 50.57 & 97.18 \\
\hline
  \end{tabular}
\label{tableMontpellierComparison}
\end{table}

\section{Conclusions}
\label{sectionConclusion}
This paper proposed a self-supervised learning based method for CD in multi-sensor bi-temporal images where one of the image is acquired by
optical sensor and the other one is captured by SAR sensor.  The proposed method effectively utilizes several concepts from the self-supervised learning, e.g., deep clustering, 
Siamese network, multiple view, and contrastive learning and operates under severe constraints, i.e., nothing except the target scene is used and no labeled data or additional unlabeled image is used.
Despite strong difference in the input modalities and operating under stringent constraints, it can identify a large fraction of the changed pixels. Comparisons with the existing methods
working under unsupervised scenario show that the proposed method brings significant improvement, especially when the target scene is large. Potential improvement of the proposed method may be achieved by 
prior learning of clusters on the unrelated domains/sensors and transferring them to target sensors on the fly \cite{menapace2020learning}. Additionally, our future work will focus on
extending the method to other application domains, e.g., comparison of biomedical images.

\section*{Acknowledgement}
The work is jointly supported by the European Research Council (ERC) under the European Union's Horizon 2020 research and innovation programme (grant agreement No. [ERC-2016-StG-714087], Acronym: \textit{So2Sat}), by the Helmholtz Association
through the Framework of Helmholtz AI [grant  number:  ZT-I-PF-5-01] - Local Unit ``Munich Unit @Aeronautics, Space and Transport (MASTr)'' and Helmholtz Excellent Professorship ``Data Science in Earth Observation - Big Data Fusion for Urban Research''(W2-W3-100) and by the German Federal Ministry of Education and Research (BMBF) in the framework of the international future AI lab "AI4EO -- Artificial Intelligence for Earth Observation: Reasoning, Uncertainties, Ethics and Beyond" (Grant number: 01DD20001).

\ifCLASSOPTIONcaptionsoff
  \newpage
\fi

\bibliographystyle{IEEEtran}
\bibliography{sarOpticalUnsupervisedCD}

\end{document}